\documentclass{article}

\usepackage[preprint]{neurips_2024}
\usepackage{microtype}
\usepackage{amsmath}           
\usepackage{enumitem}
\usepackage{graphicx}
\usepackage{bbm}
\usepackage{subfigure}
\usepackage{booktabs} 

\usepackage{amsthm}
\usepackage{amssymb}
\usepackage{mathtools}
\usepackage{mathrsfs}

\usepackage{algpseudocode}
\usepackage[vlined,linesnumbered,ruled,algo2e]{algorithm2e}
\SetKwProg{Fn}{Function}{}{}
\let\oldnl\nl
\newcommand{\nonl}{\renewcommand{\nl}{\let\nl\oldnl}}

\DeclareMathOperator{\E}{\mathbb E} 

\DeclarePairedDelimiter\autobracket{(}{)}
\newcommand{\br}[1]{\autobracket*{#1}}

\DeclarePairedDelimiter\autobrackett{[}{]}
\newcommand{\brr}[1]{\autobrackett*{#1}}

\usepackage[utf8]{inputenc} 
\usepackage[T1]{fontenc}    
\usepackage{hyperref}       
\usepackage{url}            
\usepackage{booktabs}       
\usepackage{amsfonts}       
\usepackage{nicefrac}       
\usepackage{microtype}      
\usepackage{xcolor}         

\title{Short-Long Policy Evaluation with Novel Actions}

\author{%
  Hyunji Alex Nam \\
  Department of Computer Science\\
  Stanford University\\
  \texttt{hjnam@stanford.edu} \\
  \And
  Yash Chandak \\
  Department of Computer Science\\
  Stanford University\\
  \texttt{ychandak@cs.stanford.edu} \\
  \AND
  Emma Brunskill \\
  Department of Computer Science\\
  Stanford University\\
  \texttt{ebrun@cs.stanford.edu} \\
}

\begin{document}

\maketitle

\begin{abstract}
From incorporating LLMs in education, to identifying new drugs and improving ways to charge batteries, innovators constantly try new strategies in search of better long-term outcomes for students, patients and consumers. One major bottleneck in this innovation cycle is the amount of time it takes to observe the downstream effects of a decision policy that incorporates new interventions. The key question is whether \textit{we can quickly evaluate long-term outcomes of a new decision policy without making long-term observations}. Organizations often have access to prior data about past decision policies and their outcomes, evaluated over the full horizon of interest. Motivated by this, we introduce a new setting for short-long policy evaluation for sequential decision making tasks. Our proposed methods significantly outperform prior results on simulators of HIV treatment, kidney dialysis and battery charging. We also demonstrate that our methods can be useful for applications in AI safety by quickly identifying when a new decision policy is likely to have substantially lower performance than past policies.

\end{abstract}

\section{Introduction}
We are often interested in learning and evaluating the performance of context-specific decision policies, whether for personalized adaptive educational software programs, individualized patient treatment, TV streaming platforms, or battery energy management. In many such settings, human designers frequently introduce new interventions ("actions") into the domain -- a designer may wish to add the option for a student to chat with a Large Language Model tutor, a researcher may wish to incorporate wellness coach phone calls, or new fall TV shows are now available. Typically when one is interested in evaluating the long-term outcome performance of a sequential decision policy with new actions, one simply evaluates it by executing it for many episodes (potentially in parallel) and recording its resulting performance. Unfortunately in many settings such on-policy evaluation requires a prohibitive amount of time to observe the outcomes of interest. For example,  education stakeholders are often interested in the impact of educational software on students' scores on a yearly national exam.  In reinforcement learning notation, this implies that the horizon length is one year! Similarly, long time scales are common in healthcare: for example, it is common to assess the impact of treatments on 5 year survival rates.  
Having to wait so long to assess the impact of a single policy creates considerable costs and slows the advancement of science.

It would be far preferable if one could execute a new decision policy incorporating new actions for a short horizon, and then use the data from that short trajectory to predict its performance over the full horizon. For example, we might use data gathered over 3 weeks of students using a new personalized policy incorporating new hint messages, and predict those students' end of year test scores.  We call this the \textit{short-long horizon policy evaluation} problem. While there is significant work on identifying short term surrogates of long horizon performance when an intervention is completed before the surrogate/s are observed (e.g.~\cite{athey2019}), there is very limited work on this setting where decision policies will persist through the full long horizon.  To our understanding the work that has considered decision policies that continue beyond the short term observations has focused on when the system is Markov and the dynamics can be fully identified in the short horizon, and then used to simulate the resulting policy value \cite{tang2022reinforcement, shi2023dynamic,tran2024icml}.
However, in many settings relying solely from the on-policy short-horizon data to estimate the long horizon value is challenging because the states visited and rewards obtained early on might differ substantially from later rewards. In other words, it may not be possible to identify the reward and dynamics model from the short horizon data. For example, while a personalized new reminder system to promote physical activity might yield improved initial responses, it might later prompt the individual to decrease their activity or drop out \citep{liao2019personalized}. When rewards are delayed till the end of the full horizon, as in the case of end of year school  assessments, there may be no observed rewards in the initial short trajectory. 

Fortunately, there often exists past data about other decision policies over the full horizon of interest, using other action sets-- students may have used an intelligent tutoring system (lacking the new hint messages) and past patients' treatment sequences (prior to a new drug) are recorded in electronic medical record systems. If such offline data covered the same action set as a new decision policy, even if there is no short-horizon data from the new target policy, one could use one of the many methods in offline batch policy evaluation \citep{precup, thomas15, jiang2016doubly, thomas2016dataefficient, liu19, uehara2022} to estimate its long-horizon performance. Unfortunately, such methods assume that the prior data covers the actions taken by the new decision policy, which means they typically perform poorly for our desired setting.\footnote{When the action space can be parameterized by a set of overlapping features, one can evaluate the value of new actions using a learned Q-function, but importance-sampling based methods will fail, and even fitted Q evaluation will have no state-action-tuples from the desired target policy.} 

\begin{figure}

\begin{minipage}[c]{0.65\textwidth}
\label{fig:prob_statement}
\centerline{\includegraphics[width=\textwidth]{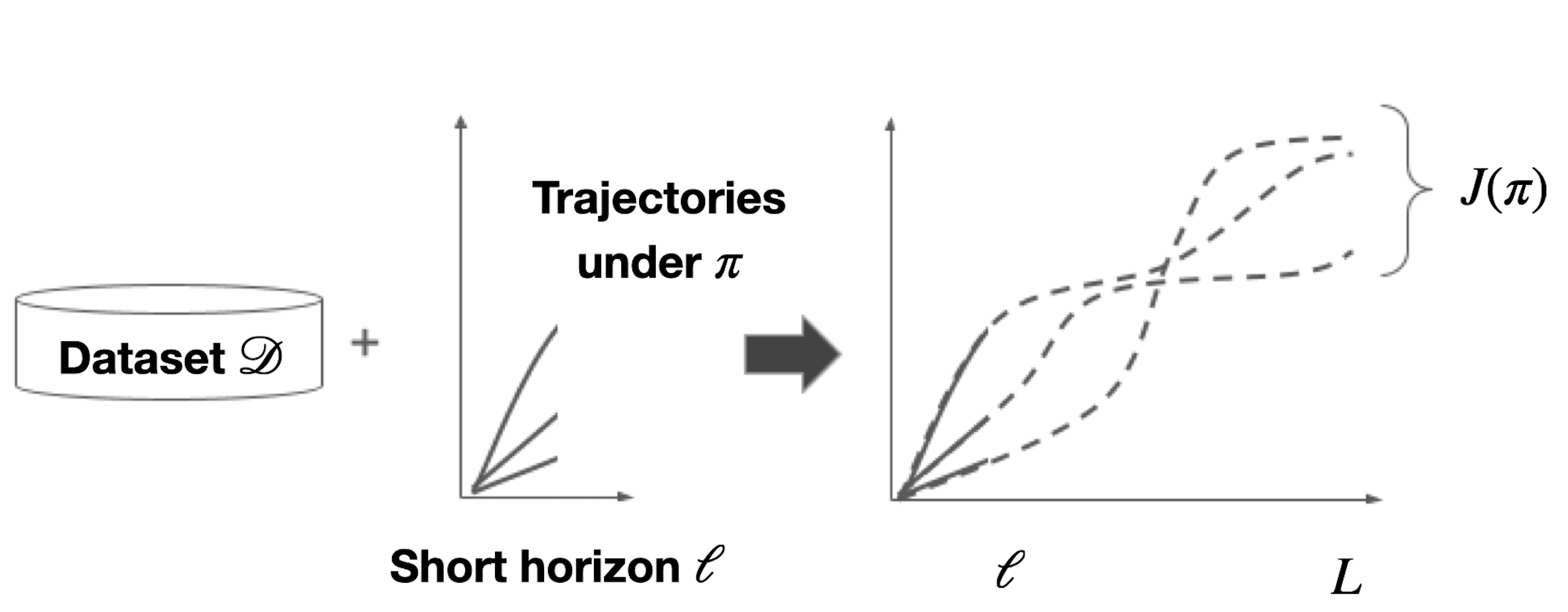}}

  \end{minipage}\hfill
  \begin{minipage}[c]{0.33\textwidth}
\caption{Short-long policy evaluation predicts the long-term outcome of a target policy with novel actions  using only short-term observations and historical off-policy data $\mathcal D$.}
  \end{minipage}
\end{figure}

In this paper, we show how we can leverage historical data about the long-horizon performance of decision policies, combined with short-horizon data of a new target decision policy with new (previously unseen) actions, to estimate the long-horizon target decision policy's value (see Figure\ref{fig:prob_statement}). We focus on the settings where the reward is a function only of the state. We first show how to frame this as an instance of supervised learning under a known distribution shift, and present the short-long evaluation of value (SLEV) algorithm. SLEV makes minimal assumptions but does not leverage the Markov dynamics structure often common in sequential decision-making tasks. To leverage such structure, we also introduce the short-long estimation of dynamics (SLED) algorithm which learns a low-dimensional parametric transformation of historical data. We demonstrate on simulators of HIV treatment, kidney dialysis and battery charging that our methods provide more accurate estimates of the long-horizon value of a target policy that uses new actions compared to alternate methods. In addition, an important task  is to be able to quickly (within a short horizon) assess if a new decision policy involving new actions may result in highly suboptimal long-term outcomes. For reasons of safety and performance, such information would allow one to quickly halt an unsafe deployment. We also show that our algorithms outperform prior baselines and can more accurately and quickly detect such scenarios.

\section{Setting}
\paragraph{Notations} For any set $\mathcal X$, let $\Delta(\mathcal X)$ be a distribution supported on the set $\mathcal X$. A sequential decision process is defined by a tuple $(\mathcal S, \mathcal A,  d_0, \mathcal P, \mathcal R)$, where $\mathcal S$ is the set of states, $\mathcal A$ is the set of actions, $d_0: \mathcal S \rightarrow [0,1]$ is the starting state distribution, $\mathcal P: \mathcal S \times \mathcal A \rightarrow \Delta(\mathcal S)$ is the transition function, and $\mathcal R:\mathcal S \rightarrow \mathbb R$ is the reward function. As rewards are often part of the model specified by the researchers themselves, we assume $\mathcal R$ is known. Let $L$ correspond to the maximum trajectory length of interacting in the environment.\footnote{For non-Markovian cases, a state $s \in \mathcal S$ comprises the history of past interactions.}

A policy $\pi: \mathcal S \rightarrow \Delta(\mathcal A)$ specifies a distribution over actions from each state. Let $S_t, A_t, R_t$ be the random variable corresponding to the state, action, and reward timestep $t$. Additionally, let $H_t^\pi \coloneqq [(S_0, A_0, R_0), ..., (S_t, A_t, R_t)]$ be the random variable corresponding to the trajectory history starting from the start state $S_0$ and then following a policy $\pi$ till time-step $t$. Further, we define the sequence of only states from the trajectory $H_t^\pi$ as $\bar H_t^\pi \coloneqq (S_0, ..., S_t)$. 

For any trajectory $H_L$, we define the return associated with that trajectory as $G(H_L) \coloneqq \sum_{t=0}^L R_t$. 
For a given $s \in \mathcal S$ and a policy $\pi$, let $v^\pi(s) \coloneqq \E_\pi\brr{\sum_{t=0}^L R_t\middle | S_0=s}$ be the value function. Let $J(\pi) \coloneqq \E_\pi\brr{G(H^\pi_L)} = \E_{S\sim d_0}\brr{v^\pi(S)}$ be the expected performance of the policy $\pi$ over the distribution of initial states over the full horizon of interest. Let $\ell < L$ be a shorter trajectory length.

\paragraph{Short-Long Policy Evaluation}

Our goal is to estimate the policy value of a new target policy $\pi'$ using only data from the first $\ell$ steps of the new policy and the historical dataset $\mathcal D$. Formally, let $\Pi_{\text{train}}$ be a set of training policies. Consider the setting where we have a full-length trajectory from each policy $\pi \in \Pi_\text{train}$:
\begin{align}
    \forall \pi \in \Pi_\text{train}\quad \br{\pi, H^\pi_\textbf{L}} &\in \mathcal D^{\text{train}},
\end{align}
and similarly data of length $\ell < L$ 
from a test policy $\pi'$ that we want to evaluate:
\begin{align}
\br{\pi', H^{\pi'}_\ell} &\in \mathcal D^{\text{test}}.
\end{align}

We allow the test policy $\pi'$ to include actions not present in the historical data $\mathcal D^{\text{train}}$. Therefore, we do not assume that $\pi(a|s) > 0$ for any policy in $\Pi_{\text{train}}$ for which $\pi'(a|s) > 0$, which makes our work different from many standard offline policy evaluation settings \cite{uehara2022}. Naturally this problem will be infeasible to solve if the new policy visits states that are not covered by $\mathcal D^{\text{train}}$. We will make this assumption more concrete in Section 4 and show when it's possible to predict $J(\pi')$.

\section{Related Work}

\textbf{Surrogate/reward proxies}. A potential approach to short-long policy evaluation is to design a short-term proxy or a surrogate that can be observed within a much shorter horizon, and use the proxy as an estimate of the long-term outcome \citep{liao2020offpolicy, linkedin, minmin_user_surrogate}. However, finding reliable short-term surrogates 
may be challenging -- for example, prior work notes that frequent short-term health behavior nudges may temporarily increase the user's physical activity, but can lead to a longer-term decrease or an early drop-out from the program \cite{aishwarya}. 
Surrogate literature in economics typically assumes surrogates completely capture any effect of the treatment and that the intervention (e.g. job training) is completed when the surrogates are measured. 
While a single feature may be insufficient, prior work for such settings has proposed using a composite surrogate index~\cite{athey2019}. Recent work has shown how a composite surrogate index can substantially reduce the necessary 
A/B testing duration in the context of an online entertainment provider~\cite{zhang2023evaluating}. In contrast to such work, we are interested
in estimating the long term outcomes of a decision policy that will continue over a long horizon, using data gathered over a shorter horizon.

\textbf{Off-policy evaluation}.
Off-policy evaluation is well-studied in RL (e.g., importance sampling-based approaches by \citep{precup, thomas15, jiang2016doubly, thomas2016dataefficient}; model-based method by \cite{liu19}; and survey of OPE methods by \cite{uehara2022}), used as an inner-loop in (model-based) RL policy optimization \citep{luo2021algorithmic, morel, mbrl_game, janner2021trust}, and also used for model predictive control in robotics \citep{ebert2018visual, williams18}. These works require that the state-actions under the test policy be supported by the data-collecting policy. Our work, on the other hand, removes this assumption and allows for actions to change between data-collection and evaluation.

\textbf{Estimating Long-Term Outcomes from Short Horizon Data}. 
There has been little prior work on estimating long-term outcomes of decision policies using short-horizon data. \citet{tang2022reinforcement} provide a method for linearly parameterized dynamics and reward model Markov decision processes with potential non-stationarity in the dynamics model. Their method assumes they have sufficient  short horizon data to identify the Markov dynamics and reward model, which is then well known to be sufficient to estimate the long horizon reward. Similarly, concurrent work by \citet{tran2024icml} assumes the dynamics and reward model can be estimated from brief short-horizon data, and then use those models to estimate the treatment effect of the value function difference between two decision policies.   
\citet{shi2023dynamic} similarly use only short-horizon data and assumes Markov structure. These authors'  construct a statistical test for comparing the difference in the (long horizon) value of two decision policies, and show this can enable them to halt experimentation earlier (when a significant result is reached) faster  than standard A/B testing. In contrast, our work considers cases where there is insufficient short horizon data to completely identify the dynamics and/or reward model in the current task, but where historical data about related tasks is available. We also consider settings which are not necessarily Markov. Other work has focused on handling (non-contextual) multi-armed bandits with delayed outcomes with partial feedback~\cite{grover2018best} also called "impatient bandits"~\cite{mcdonald2023impatient}. These papers assume some structural relationship between the shorter-term partial feedback and the delayed reward, such as a linear model~\cite{mcdonald2023impatient}, and demonstrate encouraging speed up in best-arm identification for battery charging protocol selection and hyperparameter optimization~\cite{grover2018best} and minimizing regret over podcast selection~\cite{mcdonald2023impatient}. In contrast to our work, they do not focus on contextual decision policies nor do they consider the sequential decision making setting where actions may influence later states. Related to our proposed method, ~\cite{mcdonald2023impatient} also use historical data about related actions to predict future rewards; their method uses a Gaussian prior that allows them to  capture temporal trends that are independent of arms but relate shorter term observations to final delayed rewards, in contrast to our work that can capture about state- and action-specific trajectory structure. 


Perhaps the closest to our work is that by \citet{severson} and \citet{attia_battery} on commercial lithium-ion battery charging. They show that by learning  predictive statistical models of the battery's discharge behavior from historical data, experiment time to evaluate different charging conditions can be reduced from 500 to 16 days. We will discuss and compare to their method further in the paper.

\section{Short-Long Policy Evaluation Methods}
\label{sec:proposed}

In the following section, we introduce two methods for short-long policy evaluation: One is general and can be used with or without the Markov structure, and the other leverages the Markov structure.

\subsection{Short-long estimation of value (SLEV)}\label{section:slev}
Estimating the long-horizon performance of a new decision policy with novel actions (using only short-horizon data from this policy, and long-horizon data from decision policies that use other action sets) involves prediction and addressing the action mismatch problem. Our approach leverages the following two insights:
\begin{itemize}[leftmargin=15pt]
    \item[1. ] \textbf{The prediction task:} If the short-term outcomes were correlated with the long-term outcomes and the data under the target policies have the same distribution as the data under the training policies (i.e., we simply wished to predict the performance of the policies sampled from the training set). We could reduce this to a standard supervised learning prediction problem, in which we use the short-horizon data to predict the long-horizon returns. 
    \item[2. ]  \textbf{Action mismatch problem:}  Recall that under the common setting where the reward is only a function of the state, the new policy’s performance depends only on the distribution of reached states $J(\pi) = E_{s \sim d_0} E_{s_t \sim \pi} \sum_{t=0}^L [R(s_t) | S_0 = s]$. Therefore, even if the new decision policy may incorporate new actions, we only care about the visited states to estimate its long-term value.  
\end{itemize}
The distribution over the reached states under the new policy may be different from that of the training policies. However, if the historical data has sufficient coverage over the states that are likely to be visited by the target policy, we can reduce the short-long evaluation problem to regression under a covariate shift.

The Short-Long Estimation of Value (SLEV) Algorithm (see Algorithm~\ref{alg:slev1}) presents our resulting method. Recall from our setting that $\bar H_\ell^\pi$ denotes a state-only trajectory of length $\ell$ under a  policy $\pi$. We define a function $f$ that takes as input this short-horizon trajectory $\bar H_\ell^\pi$ and predicts its long-horizon return $G(H_L^\pi)$. (Rewards may or may not be included as inputs to $f$ along with the states.) In complex settings with a finite dataset $\mathcal D^{\text{train}}$, the estimated function $f$ may not generalize well to the desired target policy $\pi'$, especially if $\pi'$ visits states with 
a low visitation frequency in the training data. Therefore, the goal is to find $f$ optimal for the (expected) test-time performance rather than the training loss.

\IncMargin{1em}
	\begin{algorithm2e}[H]
            \textbf{Input: } $\mathcal D^{\text{train}}, \mathcal D^{\text{test}}$, model class $\mathcal F$
            \\
            \textbf{Output: } $\{J(\pi')\}_{\pi' \in \Pi_{\text{test}}}$
            \\
            Split $\mathcal D^{\text{train}}$ into $\mathcal D^{\text{train-reg}}$ and $\mathcal D^{\text{train-density}}$
            \\ $\hat{w} \leftarrow \texttt{DensityRatioEstimate}(\mathcal D^{\text{train-density}}, \mathcal D^{\text{test}})$
            \Comment{$w := \frac{P_{\text{test}}}{P_{\text{train}}}$ or 1 for unweighted}
            \\
            $\hat f^* \leftarrow \arg \min_{f \in \mathcal F} \frac{1}{|\mathcal D^{\text{train-reg}}|} \sum_{(\bar H_\ell,  G) \in \mathcal D^{\text{train-reg}}} \hat w(\bar H_\ell) \left(f(\bar H_\ell) - \bar G \right)^2$ 
            \\
            \For{$\pi' \in \Pi_{\text{test}}$}
            {
            $\hat J(\pi') \leftarrow \hat f^*(\bar H_\ell')$
            }
            \Return{$\{\hat J(\pi')\}_{\pi' \in \Pi_{\text{test}}}$}
		\caption{SLEV}
		\label{alg:slev1}
	\end{algorithm2e}
	\DecMargin{1em}
 
 To do this, we  use  density ratio weighted regression \citep{ sugiyama_cv, shimodaira, cortes} to fit the function $\hat f^*$ that prioritizes states in $\mathcal D^{\text{train}}$ that are similar to those visited under the desired target policy $\pi'$.
 
\begin{equation}
\hat f^* := \underset{f}{\arg \min} \sum_{(\bar H_\ell, G) \in \mathcal D_{\text{train}}} \hat w( \bar H_\ell) \left(f( \bar H_\ell) - G \right)^2,
\end{equation} 
where $\hat w := \hat P_{\text{test}} / \hat P_{\text{train}}$ represents the (estimated) density ratio between training and testing state distributions. Eq (3) up or down-weights the training samples based on their relative likelihoods under the testing versus the training distributions. \citep{nguyen, sugiyama_breg, kptofue, zhang21} provide multiple approaches for estimating $\hat w$. 
Once $\hat f^*$ is learned, it can be used to estimate the target policy's long-term outcome $J(\pi')$ based on the short-term observation $\bar H_{\ell}'$ of length $\ell$. In our empirical results, we observe that SLEV performs well even with $\hat w= 1$ and note that the density ratio estimation step may not be necessary.

\subsection{Theoretical Analysis of SLEV Algorithm}

We can provably show the quality of the resulting estimates by the SLEV algorithm when used to estimate new policies. Let $\mathcal R_{\text{test}}$ be the test risk under $P_{\text{test}}$ (distribution of states visited by the target policies) and $\hat{\mathcal R}_{\text{train}}$ be the empirical risk of $\hat f^*$ on $(\bar H_\ell, G) \in \mathcal D^{\text{train}}, \mathcal D^{\text{train}} \sim P_{\text{train}}$. 

\textbf{Assumption 1}. \label{assumption} The training data covers the short horizon state distribution visited by target policies: 
\begin{align}
\forall \bar H_\ell \text{ s.t. } P_{\text{test}}(\bar H_\ell) > 0, \frac{P_{\text{test}}(\bar H_\ell)}{P_{\text{train}}(\bar H_\ell)} < \infty
\end{align}

With this assumption, we can bound the expected performance of $\hat f^*$ on estimates of $J(\pi'), \pi' \in \Pi_{\text{test}}$.
\newtheorem{theorem}{Theorem}[section]
\begin{theorem}
    For $\hat f^* \in \mathcal F$ (assume finite class $F = |\mathcal F|$) and a training dataset of size $n$, with probability at least $1-4\delta$,
    \begin{align}
    |\mathcal R_{\text{test}}(\hat f^*) - \hat{\mathcal R}_{\text{train}}(\hat{w} \hat f^*)| \leq M V_{\max}^2 \sqrt{\frac{\log (2F/\delta)}{2n}} + \left(\frac{M \sqrt{\log(2/\delta)}}{\sqrt{n}} + ||\hat{w} - w||_{1, P_{\text{train}}} \right) V_{\max}^2,
    \end{align} where $M := \max_{x} \{w(x), \hat w(x)\}$ and $\hat{\mathcal R}_{\text{train}}(\hat w \hat f^*) := \sum_{(x, y) \in \mathcal D^{\text{train}}} \hat w (x) (\hat f^*(x) - y)^2$.
\end{theorem}

The proof is provided in Appendix \ref{appendix:proof}. This result follows by leveraging the importance-weighted regression analysis by \cite{cortes} with \cite{kptofue}'s density ratio estimation error bound for Lipschitz density ratio functions. As expected, the performance of $\hat f^*$ for test policies depends on $M$: the (irreducible) max density ratio between train and test input state distributions, $n$: training dataset size, and $||\hat w - w||_{1, P_{\text{train}}}$: the estimation gap of $\hat w$ and true $w$.

\subsection{Short-long estimation of dynamics (SLED)}\label{section:sled}
While SLEV provides a general approach based on supervised learning, it does not leverage any Markov structure. Alternatively, we propose a dynamics model-based algorithm that leverages the Markov structure in the short and long-term data. We observe that when rewards depend only on the state, it is sufficient to simulate the states visited under a target policy $\pi'$ to estimate $J(\pi')$. A one-step state transition under $\pi'$ is:  \begin{equation}\mathbb P_1(s_{t+1}|s_t; \pi') = \sum_{a_t \in \mathcal A} \pi'(s_t, a_t) \mathcal  P(s_{t+1}|s_t, a_t).
\end{equation} 
By the Markov property, this single step transition probability can be chained to get the full state distribution $\mathbb P_t(s_t|s_0; \pi')$ at any step $t \in [L]$ starting from an initial state $s_0$.  It is well known that $P_t$ and the reward model are sufficient to estimate the policy's value: $v^{\pi'}(s_0) = \sum_{t=1}^L  \mathbb E_{S_{t} \sim \mathbb P_t(.|s_0; \pi')} \left[\mathcal R(S_{t})\right]$. Indeed the limited prior work on estimating the long horizon policy value or difference in policies' values from short-horizon data has focused on learning the Markov dynamics models $\mathcal  P(s_{t+1}|s_t, a_t)$ or $\mathbb P(s_{t+1}|s_t, \pi')$ from only short horizon data~\cite{tang2022reinforcement,tran2024icml,shi2023dynamic}.

However, learning the action-dependent dynamics model $\mathcal  P(s_{t+1}|s_t, a_t)$ or policy-dependent  dynamics model $\mathbb P(s_{t+1}|s_t, \pi')$ can be challenging given only short horizon data-- it may take many steps for the model to be identified. Using historical data to learn these models does not work when the new target decision policy $\pi'$ incorporates new actions that are not part of this historical data.

To tackle this, we draw inspiration from LoRA \citep{hu2021lora}; efficient tuning of LLMs \citep{wang2022adamix, zhang2023composing}; transfer learning through low-dim adapters \citep{oreshkin_metalearning}; and similar ideas in robotics \citep{schramm_robotics, tanaka_robotics}.

We estimate the \textit{policy-dependent} dynamics model of the new policy, $\mathbb P(s_{t+1}|s_t, \pi')$, using few-shot fine-tuning when the dynamics model can be decomposed into a  base model and a low-dim policy-specific adapter. The base dynamics model is policy-independent 
\begin{equation}
s_{t+1} = f_{\text{train}}(s_t; \theta_{\text{train}})
\end{equation}
and is trained on the historical training data $\mathcal D^{\text{train}}$ to minimize the mean squared loss\footnote{We focus on unimodal dynamics models here but our ideas could be extended to multi-modal transition settings.}. We 
take the learned $\theta_{\text{train}}$ and then for a given target policy $\pi'$, we learn a $policy-dependent$ low-dim $\lambda$ transformation of the base dynamics model: 
\begin{equation}
s_{t+1} = f_{\text{test}}^{\pi'}(s_t; \theta_{\text{train} } \circ \lambda))
\end{equation}
Note that this approach can be used even when new actions are incorporated by the target policy $\pi'$ because the base and transformed policies $f$ do not condition on the actions. 

When $\dim(\lambda) << \dim(\theta)$, learning $\lambda$ can be done with a few-shot on-policy examples from the short horizon data, which is much easier, compared to learning $\theta_{\text{train}}$ or $f_{\text{test}}^{\pi'}$ directly without any decomposition. Intuitively, we assume that the dynamics model for the new target policy is similar to the dynamics policy of prior policies, such that a low dimensional parameterization is sufficient to capture the new dynamics. We will shortly demonstrate this is a reasonable assumption in some realistic simulations. 

Finally, to estimate the policy outcome $\hat J(\pi')$, we simulate the next unobserved state using $f_{\text{test}}^{\pi'}$ and compute its reward given the predicted state. Similarly all future states are predicted auto-regressively using $f_{\text{test}}^{\pi'}$.

 Algorithm ~\ref{alg:sled} shows our Short-Long Estimation of Dynamics (SLED) Algorithm.

\IncMargin{1em}
\begin{algorithm2e}[H]
    \textbf{Input: } $\mathcal D^{\text{train}}, \mathcal D^{\text{test}}$, model classes $\Theta, \Omega$ \\
    \textbf{Output: } $\{\hat J(\pi') \}_{\pi' \in \Pi_{\text{test}}}$ \\
    $\hat \theta^* \leftarrow \arg \min_{\theta \in \Theta} \sum_{(s_t, s_{t+1}) \in \mathcal D^{\text{train}}} \left(f_{\text{train}}(s_t; \theta) - s_{t+1} \right)^2$ \;
    \For{$\pi' \in \Pi_{\text{test}}$}{
        $\hat \lambda^* \leftarrow \arg \min_{\lambda \in \Lambda} \sum_{(s_t, s_{t+1}) \in \mathcal D^{\text{test}}(\pi')} \left(f^{\pi'}_{\text{test}}(s_t; \hat \theta^* \circ \lambda) - s_{t+1} \right)^2$ \tcp{Fix $\hat \theta^*$ from line (3)}
        $\hat J(\pi') \leftarrow \texttt{CalcReturns}(\hat \lambda^* \circ \hat \theta^*, \mathcal D^{\text{test}}(\pi'))$ \;
    }
    \Return{$\{\hat J(\pi')\}_{\pi' \in \Pi_{\text{test}}}$}
    \caption{SLED}
    \label{alg:sled}
\end{algorithm2e}
\DecMargin{1em}

\section{Experiments}
 Our primary motivation for this work is to explore the possibility of producing accurate estimate of long-horizon performance for new decision policies incorporating new actions, using only short-horizon data from those policies. We evaluate our algorithms on the simulators of HIV treatment \cite{ernst}, Kidney dialysis \cite{kidney}, and battery charging \citep{severson, attia_battery} that typically take a long time, often months, to evaluate in the real world. We focus our experiments around the following questions:
\begin{enumerate}
\item Can we predict long-term policy values with SLEV?
\item Can SLEV quickly (within a very short horizon) detect safety risks (i.e., when a new decision policy is likely to have substantially lower performance than historical policies)?
\item Does leveraging the Markov assumption and the lower-rank decomposition in the SLED algorithm yield benefits over the SLEV algorithm? 
\end{enumerate}

\subsection{Domains}
To answer these questions, in each domain we construct a set of "target" decision policies for which evaluation is done, and a set of training data from a different set of behavioral policies. Full experimental details, when omitted, are provided in the Appendix \ref{appendix:experiments}.
 
\paragraph{HIV treatment} 
\begin{itemize}
\item \citep{ernst} uses a 6-dim vector of continuous values to represent a patient's state (we consider 250 initial ``unhealthy" states \cite{hiv_states}). 4 actions are possible based on two types of drugs: Reverse Transcriptase Inhibitors (RTI) and Protease Inhibitors (PI). Actions are represented as 2-dim vectors, each value corresponding to the fixed dosage per drug (i.e., drugs can either be assigned by a fixed amount or not given). The original implementation by \cite{liu19} penalizes RTI \& PI assignments as part of the reward, but we only include the state-dependent part modeling the body's response to HIV viruses.
\item Full horizon length $L$ is 200 steps, each representing a 5 day period of real-time.
\item 70 training policies based on 3 actions: RTI, PI \& both treatments. 40 new policies including the previous 3 treatments and \textit{no drug} as a new option.
\end{itemize}

\paragraph{Kidney dialysis}
\begin{itemize}
    \item \cite{kidney} represents a patient with a 5-dim continuous vector. Actions are scalar amounts of darbepoetin alfa prescribed to the patient to keep their hemoglobin (Hb) level healthy. Reward is determined by the past and current Hb concentration. 100 initial Hb states are considered in our experiments. 
    \item Full horizon length is 30, each representing one month. 
    \item Two different types of dosage controllers are available: discrete and continuous. The discrete controller discretizes the action space between [0, 1] into $n$ bins and selects among the $n$ choices; and the continuous controller can select any continuous-valued actions, particularly including those larger than 1. Each policy takes a random action with $\epsilon$ probability and otherwise takes the action suggested by the controller. 200 training policies are made based on the discrete controller, and 40 new policies based on the continuous controller are evaluated. To account for the policy's stochasticity, each policy is run 30 times from a given initial state, and $J(\pi)$ is taken as the average returns. The input to the SLEV algorithm is a flattened vector of the states and rewards from the 30 policy roll-outs.
\end{itemize}

\paragraph{Battery charging} 
\begin{itemize}
    \item A charging policy determines the conditions under which a commercial battery should be charged, and the policy's optimality is evaluated at the discharge time based on the battery's discharge cycle length \cite{severson}. Full horizon lengths may vary, but in general evaluating a battery and making policy improvements can take from months to years \citep{severson, attia_battery}.
    \item We use the dataset and code by \cite{severson} to pre-process the battery data. This dataset includes 41 batteries for training and 83 for evaluation. \cite{severson} evaluates their proposed classifier, using the data from the first 5 cycles, to classify each battery into a ``low" or a ``high" life group based on the threshold at 550 cycles. 
\end{itemize}

\subsection{Baseline Algorithms}
See \ref{appendix:baselines} for the full discussion of different baseline methods. We compare our proposed methods to existing policy evaluation approaches (Fitted Q-evaluation and online dynamics modeling) and two linear extrapolation baselines. We also include the off-policy mean as a simple predictor, which assumes the future performance of a new decision policy will be the same as the average performance of the past policies.
\begin{itemize}
    \item Fitted Q-evaluation (FQE) \cite{le2019batch} is a model-free value estimation method that does not require on-policy samples (a batch version of fitted Q-iteration by \cite{ernst2005}). We implement FQE with function approximation where states and actions are concatenated as inputs to $Q_{\theta}(s,a)$. Specifically we encode the action either as a scalar value (for Kidney dialysis) or as a 2-dim vector (for the HIV simulator) and append it to the continuous-valued state vector as the input to the Q function approximator. We selected FQE as a representative offpolicy policy evaluation method because of its flexibility with the function approximation and well-studied performance by existing works \cite{xie2020q}.
    \item Online dynamics model: We fit a dynamics model based on the short-horizon online data and use this as a simulator to evaluate the target policy's future states. The dynamics model outputs the next state given the current state and the action pair, and can auto-regressively output the full sequence of future states without running them in the actual simulator \citep{ebert2018visual, williams18}. This approach is similar to recent work that uses only short-horizon data to do long horizon policy evaluation~\cite{tang2022reinforcement} or computation of the difference between the long horizon value of two policies~\cite{shi2023dynamic,tran2024icml}. 
    \item Linear extrapolation baselines: We note that several of the health simulators in our experiments prioritize shifting the patient into a stable range of healthy states. In such settings, linear extrapolation of the rewards observed in a short horizon may be appropriate for predicting the returns (see examples \ref{fig:oracle_reward_examples}, \ref{fig:hiv_ex1}, \ref{fig:kidney_ex1}). In particular, we consider \textit{mean reward extrapolation} (averages over the short horizon reward and multiplies by the full horizon length) and \textit{last reward extrapolation} (uses the last reward and multiplies by the full horizon length). Our mean reward extrapolation is somewhat similar to \citet{tran2024icml}'s "naive" baseline in which they use the 1-step average treatment effect difference between two decision policies and multiply it by the length of the treatment duration.  However, as we will show in the results, these methods are susceptible to the choice of the short-horizon length used for observation, and do not perform consistently across domains. 
\end{itemize}

\subsection{Results}

\subsubsection{Can we predict long-term policy values with SLEV?} 
\begin{figure}

\begin{minipage}[c]{0.6\textwidth}

\centerline{\includegraphics[width=\textwidth]{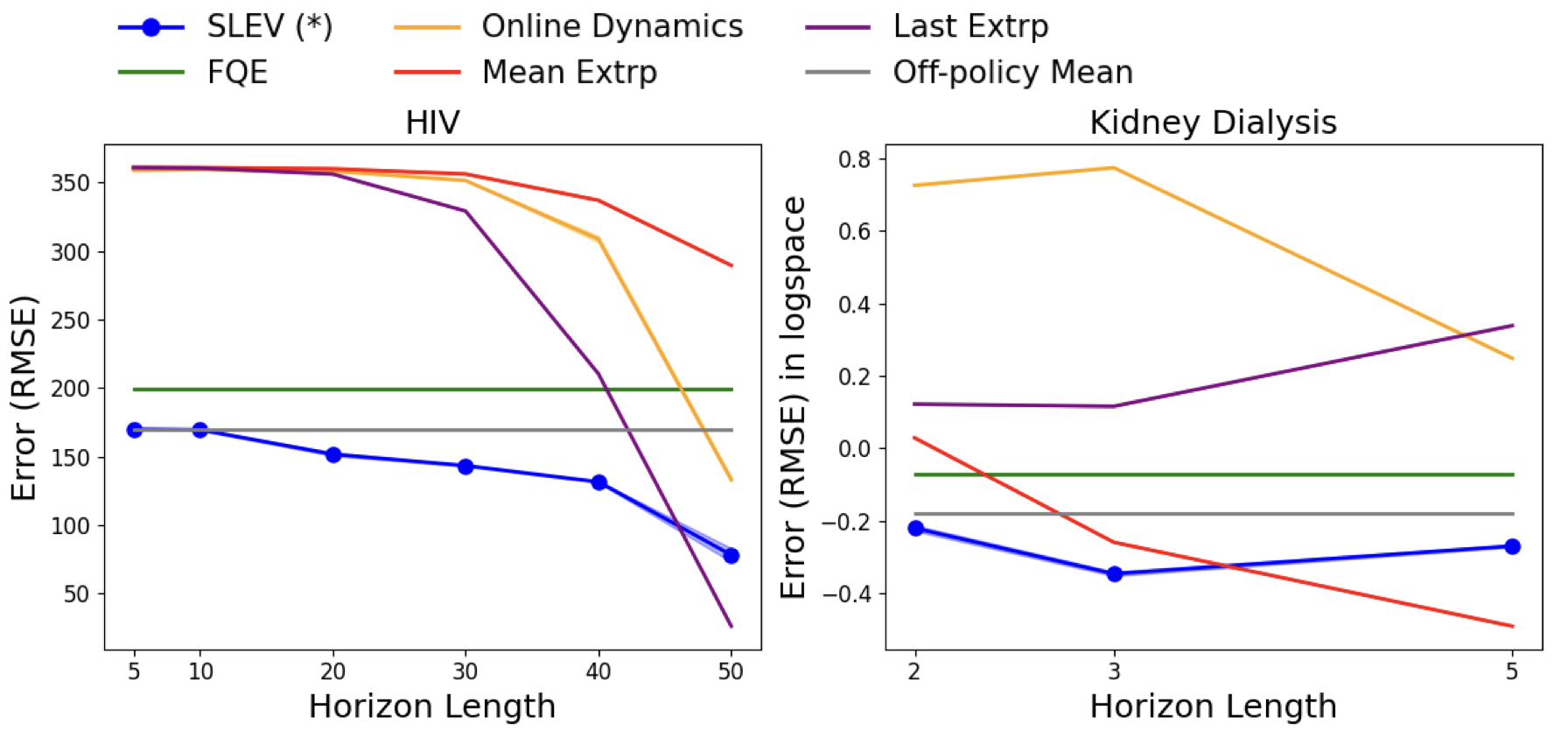}}
\caption{Comparison of methods on HIV and kidney dialysis simulators. The horizon length of 20 corresponds to 10\% of the full horizon length in the HIV treatment; 3 corresponds to 10\% in the Kidney domain. Shades show standard deviation of errors from 3 seeds.}
\label{fig:main_regression}
  \end{minipage}\hfill
  \begin{minipage}[c]{0.35\textwidth}
\centerline{\includegraphics[width=1.15\textwidth]{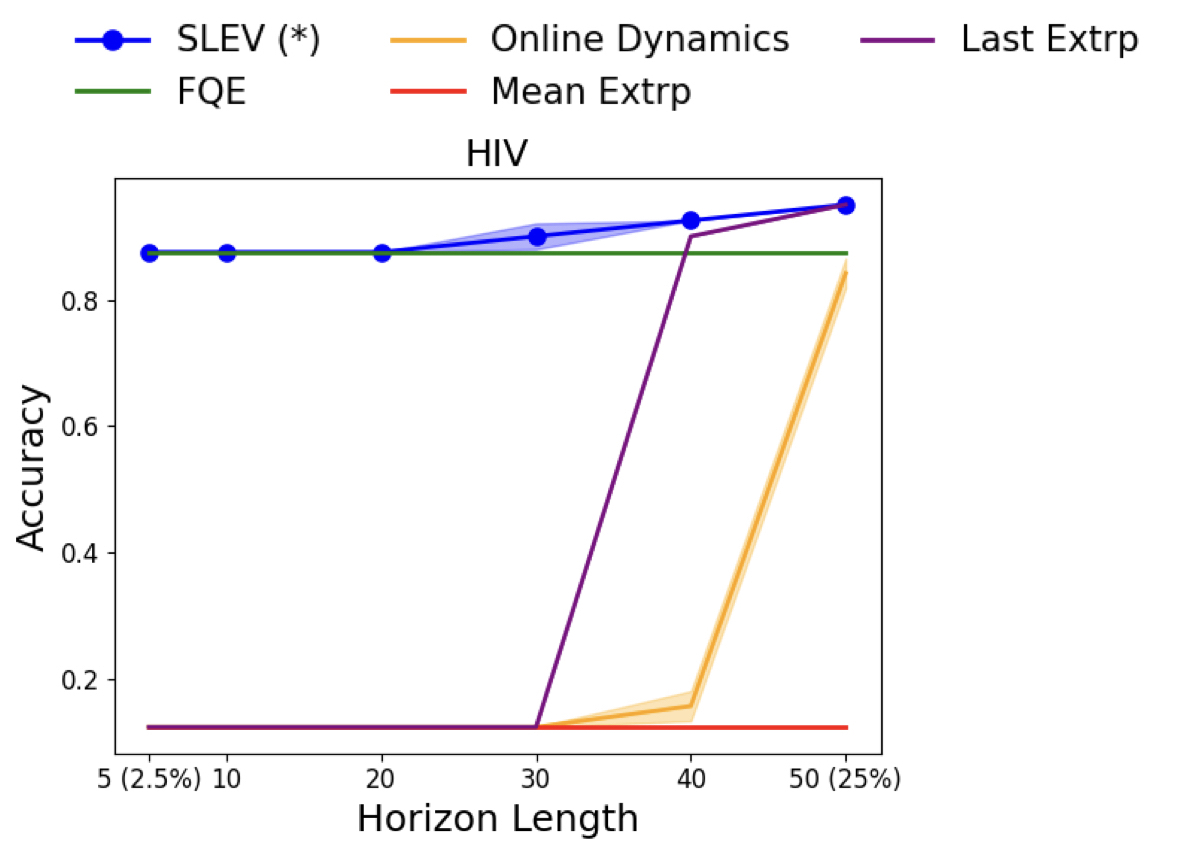}}
\caption{Comparison of different methods' accuracy of identifying safe versus unsafe policies over varying horizon lengths. Shades show standard dev from 3 seeds.}
\label{fig:safety}
  \end{minipage}\hfill
\end{figure}

Fig. \ref{fig:main_regression} shows policy prediction accuracy (in terms of RMSE) as the observed on-policy horizon length increases from 2.5\% of the full horizon to 25\% for the HIV and Kidney dialysis simulations. 

In both domains, SLEV outperforms the baselines at small horizon lengths. All methods generally improve
 given a longer input horizon, except for FQE and the off-policy-mean which do not use on-policy data. Not surprisingly, FQE, even with function approximation, generalizes poorly to the target policies incorporating the new actions as the action overlap assumption is violated.

Though in these settings linear extrapolations of short-term rewards are quite effective, this need not always be the case. First, note that the performance of the linear baselines vary widely across the observed horizon lengths and across the domains. For example, in Fig. \ref{fig:main_regression}, the last reward extrapolation (purple) performs poorly in the Kidney simulator but performs very well in the HIV simulator at $\ell = 50$. Moreover, short-term rewards can be unreliable estimates of the future outcomes. For space, we defer these examples to the Appendix (Figs  \ref{fig:kidney_ex1} and \ref{fig:hiv_ex1}). In contrast, SLEV can learn complex, and importantly, \textit{non-linear}, interactions between short and long-term effects from the historical data and use them to better evaluate new policies consistently.

\subsubsection{Can SLEV quickly detect suboptimal policies?}
When policies can lead to very different long-term outcomes, it would be highly beneficial to be able to quickly identify suboptimal policies from a short horizon. In particular, when the new policies being tested fall below a certain threshold set by the historical data, we would want to be able to mark such policies as unsafe and only continue running those that pass some initial safety check. 

Motivated by this, we evaluate SLEV's capability of quickly detecting suboptimal HIV treatment policies. As the long-term returns by the policies range widely from 10 to over 300, we set the ``unsafety" threshold at 100 which is the value of the bottom 10\% of the off-policy data. According to this threshold, 5 out of 50 policies should be flagged as unsafe based on their true long-term policy values. 

Fig. \ref{fig:safety} compares the accuracy of different methods at this task. We ignore the off-policy-mean predictor as it would give the same estimate for all new policies and not be able to distinguish between good and bad outcomes. Surprisingly, unlike in the estimation task, FQE also gives reasonably accurate detections. We hope this suggests FQE, or other offline batch evaluation methods with sufficient function approximation, could be further explored for early safety detection purposes even though their theoretical guarantees may not directly extend to such settings. Similarly, a potentially meaningful use-case of the SLEV algorithm is using it for safe deployment of new policies by helping decision-makers identify which policies are low or high risk to continue pursuing.

\subsubsection{Does leveraging the Markov assumption in the SLED algorithm yield benefits over SLEV?}
As suggested by Theorem 4.1, when the training and testing distributions change significantly, the max density ratio would be very large, leading to poor test-time generalization of the SLEV algorithm. In the HIV treatment and Kidney dialysis simulators, we observe that SLEV implemented with $\hat w = 1$ can still handle the underlying distribution shift well. However, as we still expect SLEV to suffer from a more drastic distribution shift, we investigate this problem with the battery dataset. 

First, we note that the prior work by \citep{severson, attia_battery} can be viewed as a special instance of the SLEV algorithm, where instead of using the state sequences $\bar H_\ell^\pi$ under $\pi$ directly, they do manual feature engineering to obtain a 1-dimensional feature  $\phi(H_\ell^{\pi})$ 
 and train a logistic classifier, $f: \phi(\pi) \mapsto J(\pi)$ on the transformed input with $J(\pi)$ being the battery's life expectancy. In particular, they use $\ell = 5$ and the feature is based on the discharge data from the battery's cycles 4 and 5. 
 
 To evaluate the effect of the distribution shift, we compare the classifier's performance when trained with the full dataset ($n= 41$) versus a smaller subset ($n = 24$) only including the batteries with life time shorter than 700 cycles. As expected, Table \ref{table:main_table1} shows that SLEV generalizes poorly when the training dataset is skewed (also the Appendix Fig.\ref{fig:battery_failure} shows how the function estimation may be affected, thereby making the resulting generalization poor). 
 
To solve this, we leverage the structure in the battery data that makes our second algorithm applicable. Following the first step of the SLED algorithm \ref{alg:sled}, we estimate the base function that can best capture all of the training batteries' discharge capacity curves. Given this base function, we learn a low-dim transformation for each new charging policy. In particular, we use the observations made from the first 5 discharge cycles of the battery, charged using that policy, to estimate this low-dim parameter. The Appendix \ref{appendix:sled} describes the model fitting process. Table \ref{table:main_table1} shows that SLED makes consistently good classifications even under the severe distribution shift. SLED may be preferred in certain Markov settings where the dynamics model for each policy can be decomposed for a few-shot adaptation using online samples, and therefore, learned sample-efficiently.

\begin{table}

\caption{Classification accuracy when the training data is biased. Reported from 5 seed runs.}
\begin{center}
\begin{sc}
\begin{tabular}{lcccr}
\toprule
 & SLEV & SLED \\ 
\midrule 
$\mathcal D_{\text{train}}$ &  86.7 $\pm 0.0$ \%  & 74.6 $\pm 0.0$ \% \\
$\mathcal D_{\text{filtered}}$ & 26.5 $\pm 0.0$ \% & 74.6 $\pm 0.0$ \% \\
\bottomrule
\end{tabular}
\end{sc}
\end{center}
\label{table:main_table1}
\end{table}

\section{Discussion and Conclusion}
While offline policy evaluation has received significant attention, and online policy evaluation can easily be done with Monte Carlo rollouts, such methods assume that there exists relevant data from the full horizon of interest. In this paper, we introduce the novel problem setting of estimating the long-horizon performance of a decision policy that incorporates new actions, after only executing that policy for a short horizon. This task frequently arises in applications ranging from education to healthcare where the timescales of interest can range from many months to years, and stakeholders are eager to quickly understand if new decision policies with novel interventions may yield benefit. In the common setting where reward is only a function of the state, we present two algorithms for tackling this challenge, and show that they can yield much more accurate estimates of long-horizon target policy performance than prior methods. 

One interesting area for future work is to perform short-long policy selection or optimization, such as by creating confidence intervals over the resulting target policy estimates. Another interesting direction would be to combine these ideas with action embeddings in a way that allows one to better represent the affordances of new actions and generalize from historical data. 

We think this novel problem setting is an important direction for future work, given the number of application areas with long-time horizons, and the fast pace of development of new actions/interventions that can be incorporated into decision policies. We view our work as one encouraging first step in this direction.

As with any other algorithmic decision making, the assumptions made by the proposed algorithms should be considered carefully before deploying them in any real-world scenarios, especially in high-stake decision making settings where the outcomes of interest can have a significant impact on many students, patients, and users.

\bibliography{arxiv/arxiv_bib}
\bibliographystyle{abbrvnat}

\clearpage 
\appendix
\section{Experiment Details}
\label{appendix:experiments}

\subsection{Domains}
Here we describe additional details about the simulators including: state normalization, initial state distributions, and training \& testing policy generation. Experiment code is attached as a zip file and will be made public after the anonymous review.

\paragraph{HIV treatment} The patient state is represented by a 6-dimensional vector of values in the range between $[0, 10^8]$, which we divide by $10^5$. Each action is a 2-dimensional vector: [0.7, 0], [0, 0.3], [0.7, 0.3], [0, 0] representing the fixed dosage and whether RTI or PI is assigned. Per-step reward ranges between $[-10^{300}, 10^{300}]$ and is divided by $10^8$. $J(\pi)$ is the cumulative sum of per-step rewards over 200 time steps in the simulator environment. The initial patient state is $[163573., 5., 11945., 46., 63919., 24.]$ (``unhealthy" defined by \cite{hiv_states}) and we generate 250 initial patient states by perturbing this default state vector with rate 0.6: \texttt{default state * np.random.uniform(low=-rate, high=rate, size=6)}.

We generate 70 training policies and 40 testing policies. Each policy is learned by running fitted Q-iteration (FQI) and different in two ways: the discount factor used during policy training, and the number of FQI iterations. Generally, fewer FQI iterations results in low-value policies and larger number result in high-value policies. We generate training policies by running FQI on 3 action sets: \{RTI, PI, both treatment\} $\times$ discount factors of \{0.7, 0.75, 0.8, 0.85, 0.9, 0.95, 0.98\} $\times$ iterations $[10]$. The test set is based on the augmented action set of \{RTI, PI, both, no treatment\}. 40 evaluation policies are generated with discount factors \{0.5, 0.8, 0.9, 0.98\} also with iterations $1, 2,..., 10$ of fitted Q iteration. Note generally the more the iteration the better the policy performs. We design our policies to cover a wide range of long-term behaviors by varying the discount factors and the number of iterations.

\paragraph{Kidney dialysis} The patient state ranges between $[-500, 500]$ and per-step reward between $[0, 10]$. States are divided by 500 and per-step reward by 10. $J(\pi)$ is the cumulative sum of per-step rewards over 30 steps representing 30 months of kidney dialysis treatment. We generate 100 initial patient states by sampling the hemoglobin level uniformly at random from $[8, 13]$.

We generate 200 training policies, and 40 testing policies. Each policy is rolled out 30 times because the policies are $\epsilon$-greedy (choosing a random action with $\epsilon$ probability at each time step). The same set of 100 initial patient states are used across all policies. 

In order to ensure training and testing policies have non-overlapping actions, we base them on two different controller types for determining darbepoetin alfa dosage: discrete and continuous implemented by \cite{kidney}. Specifically, we use the discrete controller as the base model for training policies, and the continuous controller for evaluation policies. The discrete controller divides up the action space between [0, 1] into $n$
discrete bins and chooses from the discrete action set of size $n$. The continuous controller in/decreases the dosage amount by 25\% each time based on the body’s response. 

The key difference between the discrete and continuous controller actions is that the discrete actions are bounded between [0, 1], while the continuous actions are unbounded. Policies are stochastic with $\epsilon$ determining the amount of stochasticity in the actions. With $\epsilon$ probability, the training policies choose a random action $\sim \text{Unif}([n])$, and evaluation policies choose a real-valued action $\sim \text{Unif}([0, 1])$. The training policies are different in terms of $n$ and $\epsilon$; $n \sim \text{Unif}(\{5, ..., 20\})$ and $\epsilon \sim \text{Unif}((0.05, 0.5))$. Testing policies are sampled from the same $\epsilon$ distribution as the training policies.

\begin{figure}[ht]
\vskip 0.2in
\begin{center}
\centerline{\includegraphics[scale=0.35]{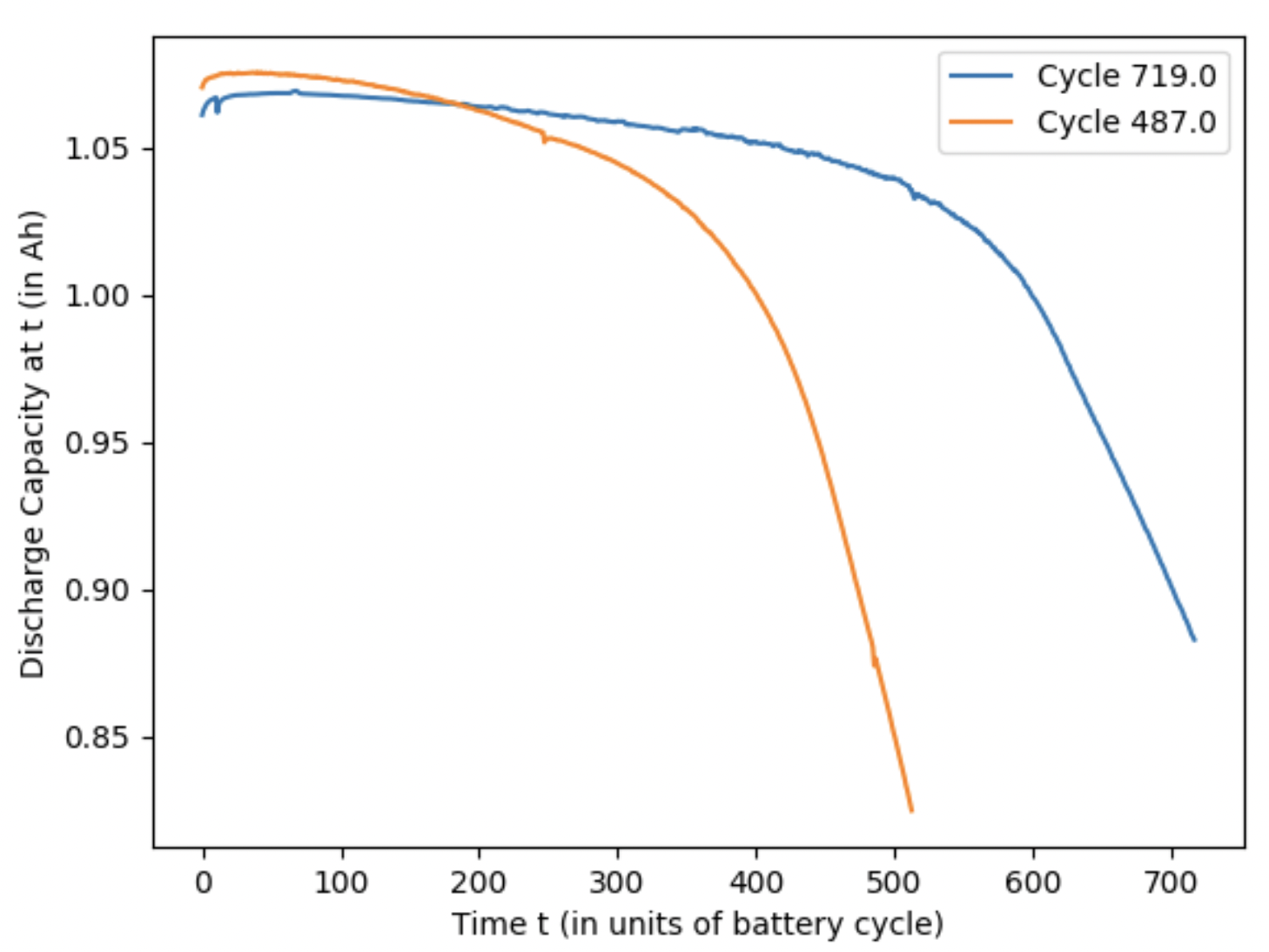}}
\caption{Two different battery cell curves, one with the lifetime of 719 cycles, and the other one with the lifetime of 487 cycles.}
\label{fig:sup_battery_sample}
\end{center}
\vskip -0.2in
\end{figure}

\paragraph{Battery charging} Each battery charging policy yields a unique discharge capacity curve, which is shown in Fig. \ref{fig:sup_battery_sample}. \cite{severson} generated 41 training samples and 83 samples for evaluation. ``Data generation" part of their paper details how the battery charging policies differ for interested readers. In short, they determine charging rates, and the batteries are evaluated on their performance at the discharge-time. We use the same training and test data as the original work by \cite{severson}.

\cite{severson} featurizes each battery into a 1-dimensional value based on the difference of the discharge voltage curves measured at the cycles 4 and 5. In reproducing their result as the baseline comparison to the SLED algorithm, we use the code by \cite{severson} to featurize the input and build a logistic regression using \texttt{sklearn LogisticRegressionCV with cv=10}.

\subsection{Algorithm Implementation}
\IncMargin{1em}
\begin{algorithm2e}[H]
    \textbf{Input: } $\mathcal D^{\text{train}}$ collected by running policies $\in \Pi_\text{train}$, hyper-parameter set $\Lambda$, and $\mathcal D^{\text{test}}$ \\
    \textbf{Output: } $\{J(\pi')\}_{\pi' \in \Pi_{\text{test}}}$ \\
    Split $\mathcal D^{\text{train}}$ into $\mathcal D^{\text{train-regression}}$ and $\mathcal D^{\text{train-density}}$ \;
    $\hat{w} \leftarrow \texttt{DensityRatioEstimate}(\mathcal D^{\text{train-density}}, \mathcal D^{\text{test}})$ \tcp*[f]{$w := \frac{P_{\text{test}}}{P_{\text{train}}}$}
    Split $\mathcal D^{\text{train-regression}}$ into $\{\mathcal D_i\}_{i=1}^k$ based on $\Pi_\text{train}$ \tcp{$k$-fold cross-validation}

    \vspace{5pt}
    \nonl \textcolor[rgb]{0.5,0.5,0.5}{\# For loops for both $i$ and $\lambda$ can be parallelized.}

    \For{$i \in [1,\hdots, k]$}{
        \For{$\lambda \in \Lambda$}{
            ${\theta_{i, \lambda}} \leftarrow \texttt{Train}(\lambda, \hat w, \mathcal D\backslash \mathcal D_i, \texttt{WeightedLoss})$ \tcp{Use any loss minimization step}
        }
    }

    $\lambda^* \leftarrow \arg\min_{\lambda \in \Lambda} \sum_{i=1}^k \texttt{WeightedLoss}(\theta_{i,\lambda}, \hat w, \mathcal D_i)$ \tcp{Best hyper-parameter} \;
    $\hat f^* := \sum_{i=1}^k \frac{1}{k} f_{\theta_i, \lambda^*}$ \tcp{Best regression model} \;
    \Return{$\{J(\pi') := \hat f^*(\mathcal H_\ell')\}_{\pi' \in \Pi_{\text{test}}}$} \;

    \SetKwProg{Fn}{Function}{:}{}
    \SetKwFunction{WeightedLoss}{WeightedLoss}
    \Fn{\WeightedLoss{$\theta, \hat w, \mathcal D$}}{
        \Return{$\frac{1}{|\mathcal D|}\sum_{(\mathcal H_\ell, \bar G) \in \mathcal D} \hat w(\mathcal H_\ell) \left(f_\theta(\mathcal H_\ell) - \bar G \right)^2$}
    }
    \caption{SLEV}
    \label{alg:full_1}  
\end{algorithm2e}
\DecMargin{1em}

For model selection over $\Lambda$, we instantiate SLEV with k-fold cross validation shown in Alg \ref{alg:full_1}. For our experiments on the simulators of HIV treatment and kidney dialysis, we observe good regression performance with $\hat w = 1$, and do not perform the additional step of \texttt{DensityRatioEstimate}. Empirically if the distribution shift between train and test states is expected to be moderate, weighted regression can be skipped. 

\paragraph{Hyperparameters for HIV} We use $k = 10$ fold cross validation with the Adam optimizer with learning rates: $\{10^{-2}, 10^{-3}, 10^{-4}\}$, and weight decay: $\{25, 10, 1, 10^{-1}\}$. 8000 gradient update steps with 100 samples per update, and early stopping is allowed if the validation losses do not improve. For the online dynamics model baseline, similarly, we use the Adam optimizer with learning rates: $\{10^{-2}, 10^{-3}, 10^{-4}\}$ and weight decays: $\{25, 10, 1, 10^{-1}, 10^{-2}, 10^{-3}\}$ for hyperparameter sweeping. For FQE, we use \texttt{ExtraTreesRegressor} from \texttt{sklearn.ensemble} as the function approximator class and vary the number of trees (25, 50, 75) in the ensemble and the number of Q function updates (1k, 2.5k, 5k). Each batch used for the update has 1000 state-action tuples randomly selected from the offpolicy dataset. We selected the best hyperparameter pair based on the performance on the held-out training set and used the same hyperparameter to evaluate FQE for all new testing policies. 

\paragraph{Model architecture for HIV} We use a multi-layer perceptron (MLP) with ReLU activation between all intermediate layers. The model layers are: $[7 * \ell, 50], [50, 25], [25, 10], [10, 1]$ with $7 * \ell$ being the input dimension of 6-dimensional state vector concatenated with the scalar reward times the observed horizon length (= $\ell$). We used the same model architecture for the online dynamics model baseline but the output is the 6-dim state vector instead of a scalar value.

\paragraph{Hyperparameters for kidney dialysis} We use $k = 10$ fold cross validation with the Adam optimizer with learning rates: $\{10^{-2}, 10^{-3}, 10^{-4}\}$, and weight decay: $\{1, 5^{-1}, 10^{-1}, 5^{-2}\}$. 8000 gradient update steps were made with 100 samples per update, and early stopping is allowed if the validation losses do not improve. Similarly for the online model baseline, we use the Adam optimizer with learning rates: $\{10^{-2}, 10^{-3}, 10^{-4}\}$, and weight decay: $\{1, 5^{-1}, 10^{-1}, 5^{-2}\}$.

For FQE, we use \texttt{ExtraTreesRegressor} from \texttt{sklearn.ensemble} as the function approximator class and vary the number of trees (25, 50, 75) in the ensemble and the number of Q function updates (1k, 2.5k, 5k). Each batch used for the update has 1000 state-action tuples randomly selected from the offpolicy dataset. We selected the best hyperparameter pair based on the performance on the held-out training set and used the same hyperparameter to evaluate FQE for all new testing policies. 

\paragraph{Model architecture for kidney dialysis} Each input to the regression model has dimension $30 * 6 * \ell$, where 30 = number of roll-outs (to account for the policy's stochasticity), 6 = state dimension + scalar reward per step, $\ell$ = observed horizon length. We have two parts to the model architecture: linear layer followed by non-linear activation layers. The linear layer takes in the input of length $6\ell$ and reduces it to $\frac{6\ell}{2}$ repeatedly for each of the 30 roll-outs. Thus outputted $30 * 3\ell$-dimensional vector is flattened into a single vector and used as inputs to the non-linear model. 

We perform model selection over two non-linear model architectures: small model has layers $[90\ell, 100], [100, 20], [20, 1]$; large model has $[90\ell, 500], [500, 300], [300, 100], [100, 20], [20, 1]$. Both models have ReLU activation between all intermediate layers. 

For the online dynamics model baseline, we use 4-layer model with the input being 5-dim states concatenated with a scalar-valued action (so a 6-dim vector) and the output as the 5-dim vector representing the next state. The model layers are: $[6, 20], [20, 10], [10, 10], [10, 5]$, with ReLU activation between all intermediate layers.

\IncMargin{1em}
\begin{algorithm2e}[H]
    \textbf{Input: } $\mathcal D^{\text{train}}, \mathcal D^{\text{test}},$ hyper-parameter sets for train and test dynamics $\Theta, \Lambda$ \\
    \textbf{Output: } $\{\hat J(\pi')\}_{\pi' \in \Pi_{\text{test}}}$ \\
    $\hat \theta^* \leftarrow \texttt{FitGlobalModel}(\Theta, \mathcal D^{\text{train}})$ \;
    \For{$\pi' \in \Pi_{\text{test}}$}{
        $\mathcal D^{\text{test}}(\pi') := \{(s_t, s_{t+1})\}_{t=0}^{\ell-1}$ \;
        Partition $\mathcal D^{\text{test}}(\pi')$ into $\{D_i\}_{i=1}^k$ \tcp{k-fold cross validation}
        \For{$i \in [k]$}{
            \For{$\lambda \in \Lambda$}{
                $f_{\lambda, i} \leftarrow \arg \min_{f \in \mathcal F_\lambda} \sum_{(s_t, s_{t+1}) \in \mathcal D^{\text{test}}(\pi') \backslash D_i} \left(f(s_t; \hat \theta^*) - s_{t+1} \right)^2$ \;
                $\mathcal L(\lambda, i) \leftarrow \sum_{(s_t, s_{t+1}) \in D_i} \left(f_{\lambda, i}(s_t; \hat \theta^*) - s_{t+1}\right)^2$ \;
            }
        }
        $\hat \lambda^* \leftarrow \arg \min_\lambda \sum_{i=1}^k \mathcal L(\lambda, i)$ \;
        $\hat J(\pi') \leftarrow \texttt{CalcReturns}(\hat \lambda^*, \hat \theta^*, \mathcal D^{\text{test}}(\pi'))$ \;
    }
    \Return{$\{\hat J(\pi')\}_{\pi' \in \Pi_{\text{test}}}$}  
    \vspace{5pt}
\SetKwFunction{FitGlobalModel}
{FitGlobalModel}
\Fn{\FitGlobalModel{$\Theta, \mathcal D^{\text{train}}$}}
   { 
   $\mathcal D^{\text{train}} := \{(s_t, s_{t+1})\}_{t=0}^L$
   \\
   Partition $\mathcal D^{\text{train}}$ into $\{D_i\}_{i=1}^k$ \Comment{k-fold cross validation} \\
   
        \For{$i \in [k]$}
        {
        \For{$\theta \in \Theta$}
            {
            $f_{\theta, i} \leftarrow \arg \min_{f \in \mathcal F_\theta} \sum_{(s_t, s_{t+1}) \in \mathcal D^{\text{train}} \backslash  D_i} \left(f(s_t) -s_{t+1} \right)^2$
            \\ 
            $\mathcal L(\theta, i) \leftarrow \sum_{(s_t, s_{t+1}) \in D_i} \left(f_{\theta, i}(s_t) - s_{t+1} \right)^2$
            }
        }
        $\hat \theta^* \leftarrow \arg \min_{\theta \in \Theta} \sum_{i=1}^k \mathcal L(\theta, i)$
        \\
        \Return{$\hat \theta^*$}
   }
  \vspace{5pt}
    \SetKwFunction{CalcReturns}{CalcReturns}
    \Fn{\CalcReturns{$\hat \lambda^*, \hat \theta^*, \mathcal D^{\text{test}}(\pi')$} }
    { 
       $\hat s_\ell \leftarrow s_\ell$ \Comment{$s_\ell$ = last time step in $\bar H_\ell(\pi')$}
       \\ 
       $\hat J(\pi') \leftarrow \sum_{t=0}^\ell r_t$
       \\
       \For{$t \in [\ell + 1, ..., L]$ }{
       $\hat s_{t} \leftarrow f(\hat s_{t-1}; \hat \theta^* \circ \hat \lambda^*)$
       \\ 
       $\hat J(\pi') \leftarrow \hat J(\pi') + \texttt{Reward}(\hat s_t)$
       }
       \Return{$\hat J(\pi')$} 
      }  
\caption{SLED}
\label{alg:sup_full_2}  
\end{algorithm2e}
\DecMargin{1em}

We present the k-fold cross validation version of SLED to choose appropriate hyper-parameters from $\Theta$ and $\Lambda$ in \ref{alg:sup_full_2}. There are two rounds of cross validation: one for learning $\hat \theta^*$ from $\mathcal D^{\text{train}}$ and the other for estimating $\hat \lambda^*$ for each testing policy $\pi' \in \Pi_{\text{test}}$ based on $\mathcal D^{\text{test}}(\pi')$. Then based on $\hat \theta^* \circ \hat \lambda^*$, SLED auto-regressively predicts the next state $\hat s_t$ for all remaining time steps $\ell+1, ..., L$. Assuming \texttt{Reward} is known given states, $\hat J(\pi')$ is computed as the sum of predicted per-step rewards.

\paragraph{SLED for battery charging} \label{appendix:sled} 

\begin{figure}[ht]
\vskip 0.2in
\begin{center}
\centerline{\includegraphics[scale=0.4]{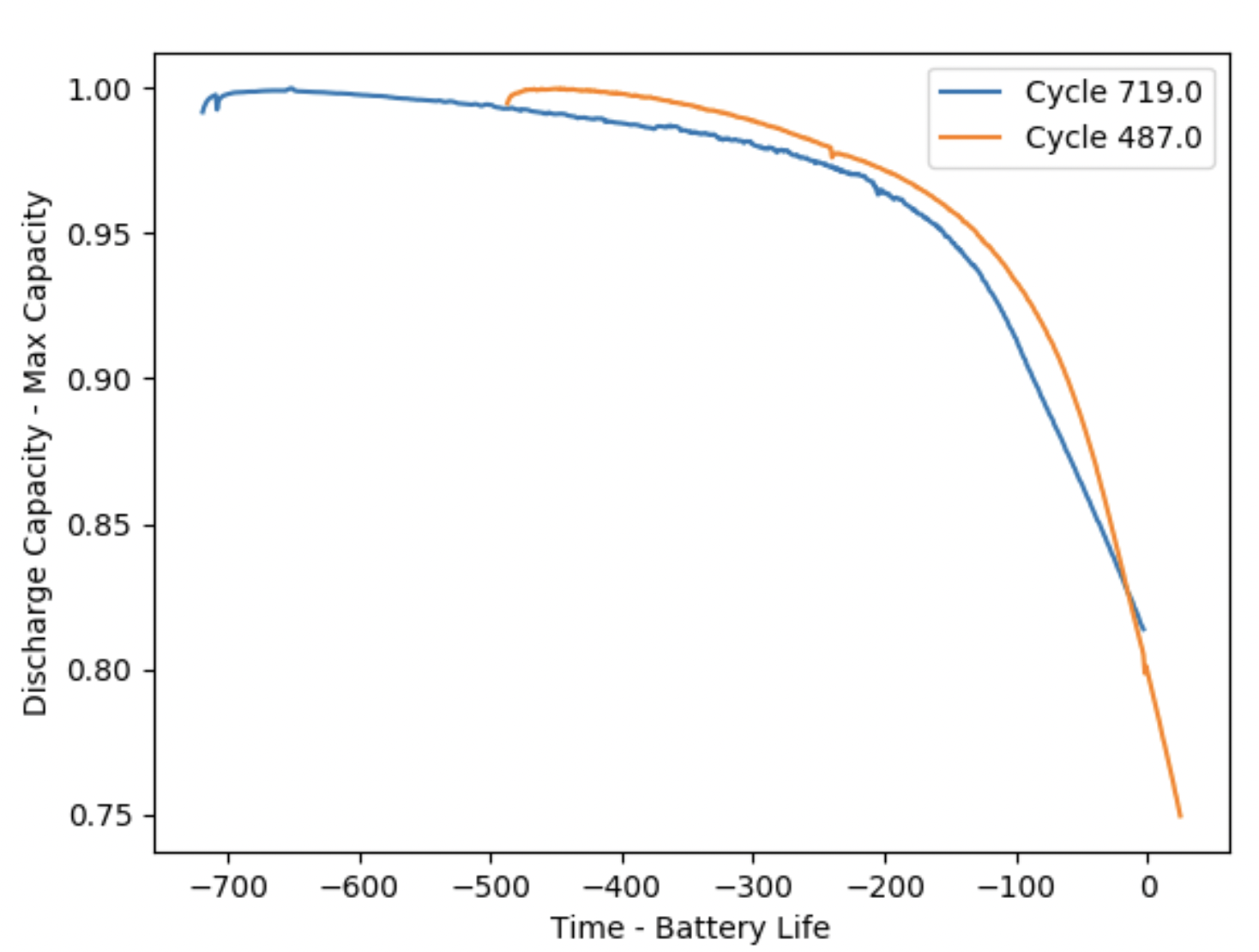}}
\caption{Battery curves, same ones from Fig. 6, become right-aligned after being shifted by their life cycles along the $x$-axis. Both curves can be explained by a global curve $f$.}
\label{fig:sup_battery_shifted}
\end{center}
\vskip -0.2in
\end{figure}

Fig \ref{fig:sup_battery_shifted} shows the shifted (right-aligned) battery curves, where the shift is determined by each battery's lifetime. We first shift all battery samples in $\mathcal D^{\text{train}}$ by their observed lifetimes and subtract $\texttt{max(discharge capacity)} -1$ from \texttt{discharge capacity(t)} for each discharge cycle $t$ in order to ensure the discharge value doesn't exceed 1. For model fitting on train data ($\hat \theta^*$), we consider 3 model classes: negative exponential, linear, and quadratic. The negative exponential model has two parameters $(a, b)$: $f(t, a, b) = \frac{1}{1+\exp(-a t + b)}$ where $t$ is the (shifted) discharge cycle number and $f(t, .)$ represents the discharge capacity from that cycle. Parameters $a, b$ are fitted on the train data. The linear model also has two parameters $(a, b)$: $f(t, a, b) = a t + b$, and quadratic has 3 parameters $(a, b, c)$: $f(t, a, b, c) = a t^2 + b t + c$. We use $k = 10$ fold cross validation on the train data to find the best model class and the best parameters $a, b$ (and $c$ if applicable). 

Once the model class and the parameters are selected based on the train data validation loss, we fit a local parameter $\hat \lambda^*$ per battery curve in the test set. During evaluation, the battery life time is unknown (in fact this is the prediction target, which we call \texttt{lfc}). We can still subtract \texttt{max(discharge capacity)} based on the observed the first $\ell$ cycles discharge capacity (since discharge capacity drops over time). Model fitting during testing ($\hat \lambda^*$) estimates \texttt{lfc} by considering two model classes: $f(t - \texttt{lfc})$ and $f(t - \texttt{lfc}) + y$. The first model assumes the same parameters $a, b, (c)$ as before and only estimates $\texttt{lfc}$ while the second model estimates $\texttt{lfc}$ along with a vertical shift in the $y$ direction. We again perform $k=10$ fold cross validation on the test battery data (unless the data points are too few then use $k = 5$ fold instead). In the battery charging domain, we do not need to auto-regressively compute future states and convert them to rewards because \texttt{lfc} directly gives the long-term outcome of interest. But in general \texttt{CalcReturns} is needed as an additional step to predict future states and rewards at $\ell+1, ..., L$. 

\subsection{Baselines}
\label{appendix:baselines}
\begin{table}
\caption{Shows whether a given method (1) requires observing on-policy samples $\tau_h^\pi$, (2) leverages historical data $\mathcal D$ to build a model prior to observing $\tau_h^\pi$}
\label{table:methods}
\vskip 0.15in
\begin{center}
\begin{small}
\begin{sc}
\resizebox{\columnwidth}{!}{%
\begin{tabular}{lcccr}
\toprule
Method & On-policy samples? & Use $\mathcal D^{\text{train}}$?  \\
\midrule
SLEV (proposed) & \checkmark & \checkmark \\
SLED (proposed) & \checkmark & \checkmark  \\
Online dynamics model & \checkmark & \text{\sffamily X} \\
Fitted Q-evaluation & \text{\sffamily X} & \checkmark  \\
Average reward extrapolation & \checkmark & \text{\sffamily X}  \\ 
Last reward extrapolation & \checkmark & \text{\sffamily X}  \\
\bottomrule
\end{tabular}%
}
\end{sc}
\end{small}
\end{center}
\vskip -0.1in
\end{table}  

We evaluate our proposed algorithms against a suite of existing OPE methods. Table \ref{table:methods} compares these methods in terms of whether they (1) require any on-policy samples from $\pi_{\text{test}}$, and (2) utilize historical data $\mathcal{D}^{\text{train}}$.

\paragraph{Offline policy evaluation:} Fitted Q-evaluation is a standard off-policy evaluation method that can estimate the return of a target policy using tuples from a (different) behavioral policy. Because the actions are non-overlapping between train and test policies, standard FQE may not generalize well to unseen actions from test policies.\footnote{But we observe in the HIV experiments that using function approximation to represent the Q function helps mitigate this issue to some degree.} We use function approximation to represent $Q^{\pi'}(s, a) = f^{\pi'}_\theta(s, a)$. \begin{equation}
\hat f^{\pi'} _\theta(s, a):= \arg \min_{f_\theta} \mathbb E_{(s, a, s') \sim \mathcal D^{\text{train}}} \left[f_\theta(s, a) - \{r(s, a) + \gamma f_\theta(s', \pi'(s'))\}\right]^2
\end{equation} where $\mathcal D^{\text{train}}$ is the dataset of tuples $(s, a, s')$ collected under a mixture of policies $\Pi_{\text{train}}$, and the evaluation policy is $\pi' \in \Pi_{\text{test}}$. Then $\hat J(\pi') := \widehat{\mathbb E}_{ s \sim d_0} [\hat f_{\theta}^{\pi'}(s, \pi'(s))]$ is the policy value estimate under the target/evaluation policy over the initial starting state distribution.

\paragraph{Online on-policy dynamics model:} We only use on-policy data in $\mathcal D^{\text{test}}$ to learn a single-step state transition model $f$ as follows, 
\begin{equation}
\arg \min_{\theta} \sum_{(s_t, a_t, s_{t+1}) \in \mathcal D^{\text{test}}} \left(f_\theta(s_t, a_t) - s_{t+1}\right)^2.
\end{equation} While model-based RL typically trains an transition dynamics model using $\mathcal D^{\text{train}}$ and tests on policies that have overlapping actions as $\mathcal D^{\text{train}}$, we only use on-policy examples in $\mathcal D^{\text{test}}$. This is because we explicitly make our experiments to violate the standard action-overlap assumption.

\begin{figure}[ht]
\vskip 0.2in
\begin{center}
\centerline{\includegraphics[scale=0.15]{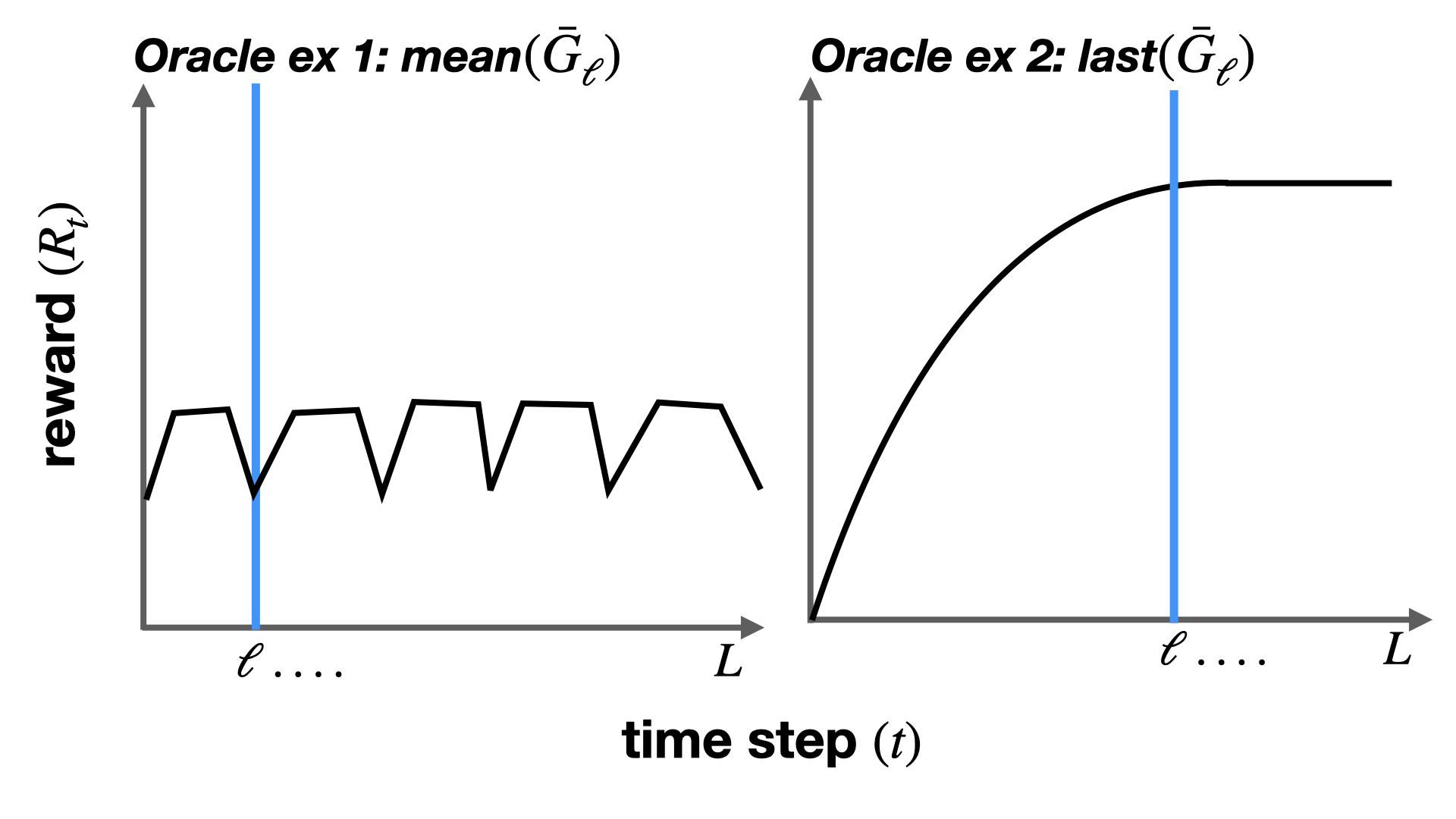}}
\caption{Left reward sequence can be best predicted by \textit{average reward extrapolation}; right reward sequence by \textit{last reward extrapolation}. If reward sequences look like the above examples, the naive heuristics achieve oracle performance, but the main drawback of these approaches is that the appropriate $\ell^*$ is unknown in most systems.}
\label{fig:oracle_reward_examples}
\end{center}
\vskip -0.2in
\end{figure}

We also implement two heuristic-based linear extrapolation approaches that show oracle performance in settings with very specific structures (see Fig\ref{fig:oracle_reward_examples} for examples). The purpose of including them in the experiments is not to encourage these methods (since they require explicit knowledge about the environments which is unavailable most of the times) but rather to show how our proposed algorithms (SLEV and SLED) compare with these approaches which assume more knowledge about the systems.

\paragraph{Average reward extrapolation:} Observe reward until some time step $\ell$ under the target policy $\pi'$. Then use the observed so far, $(R_t)_{t=0}^\ell$, to linearly extrapolate the future returns under that policy.  \begin{equation}
\hat J(\pi') := \frac{L}{\ell} \sum_{t=0}^\ell R_t.
\end{equation} We observe that this achieves oracle performance with an appropriate $\ell^*$ in Kidney dialysis treatment because patient's state dynamics undergo cycles in both (discrete and continuous) treatment regimes. 

\paragraph{Last reward extrapolation:} Similarly as above, observe rewards until $\ell$, then extrapolate the future outcome based on the last observed reward only:
\begin{equation}
\hat J(\pi') := L * R_\ell.
\end{equation} While this seems very naive, we observe that this achieves oracle performance in HIV treatment with an appropriate $\ell^*$ because the ground truth dynamics plateau after some initial phase (the exact time at which the states start to plateau vary across different policies and patient states, but as long as $\ell^*$ is sufficiently long to observe the plateauing point, ``last reward extrapolation" achieves the best prediction for $\hat J(\pi')$). However, the main drawback of using these naive heuristics is that the appropriate $\ell^*$ is unknown in most systems until the full $L$-step trajectories are observed (which we have access to for both our train and test policies in our controlled experiments).

\subsection{Training compute requirements}
Our training doesn't require any GPUs since it only requires a few MLP training for HIV and Kidney dialysis and uses \texttt{sklearn LogisticRegressionCV} and \texttt{scipy optimize} for the battery charging experiments. Model selection can be parallelized, but if run in series, can take $\sim$ 1 day for hyper-parameter search and policy evaluation. The training code is available in the Supplementary material and will be released as a github codebase after the anonymous review.  

\section{Results}
\subsection{Qualitative examples for SLEV}
For both HIV and Kidney treatment domains, we show examples where early reward signal is misaligned but SLEV can correctly identify the long-term policy value based on the short-term observation.

\begin{figure}[ht]
\vskip 0.2in
\begin{center}
\centerline{\includegraphics[scale=0.35]{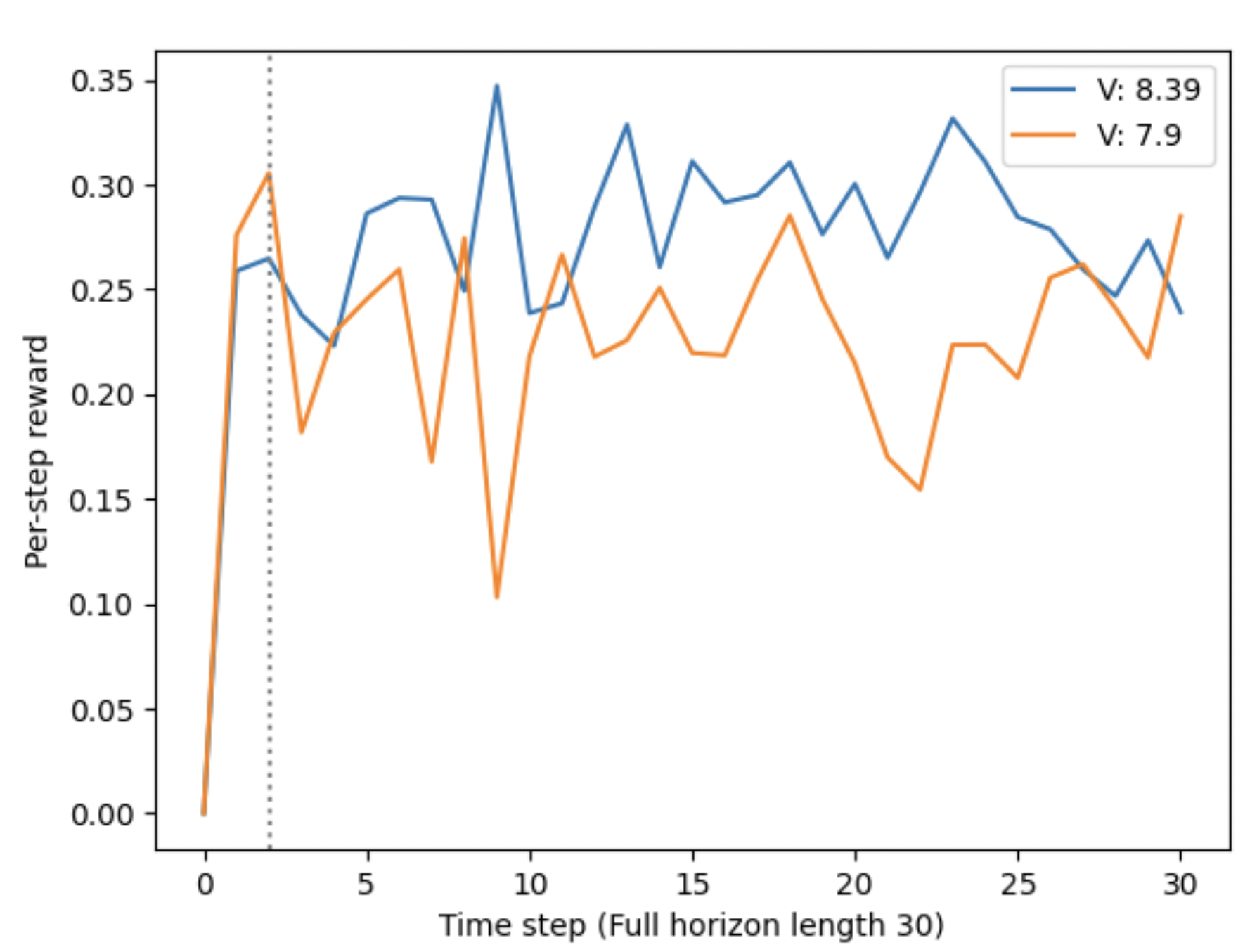}}
\caption{Per-step rewards obtained while following two different policies in Kidney dialysis treatment. The low-value policy (orange) achieves a higher initial reward mean than the high-value policy (blue). SLEV can still correctly rank these policies based on their predicted outcomes and not be misguided by their initial rewards but ranking based on the initial rewards until the dotted vertical line would yield incorrect predictions about which policy is preferred in the long run.}
\label{fig:kidney_ex1}
\end{center}
\vskip -0.2in
\end{figure}

Fig \ref{fig:kidney_ex1} from the kidney domain show two per-step reward sequences under two different policies. The blue trajectory achieves the long-term value of 8.39 while the orange trajectory achieves 7.9. The dotted gray vertical line shows where the last observation is made ($\ell = 2$). Mean reward extrapolation is misguided because the orange trajectory achieves higher rewards than the blue trajectory in the initial time steps, but SLEV still correctly predicts the value of the blue trajectory to be higher ($\hat J(\pi') = 8.41$) than that of the orange policy ($J(\pi'') = 7.86$). 

\begin{figure}[ht]
\vskip 0.2in
\begin{center}
\centerline{\includegraphics[scale=0.35]{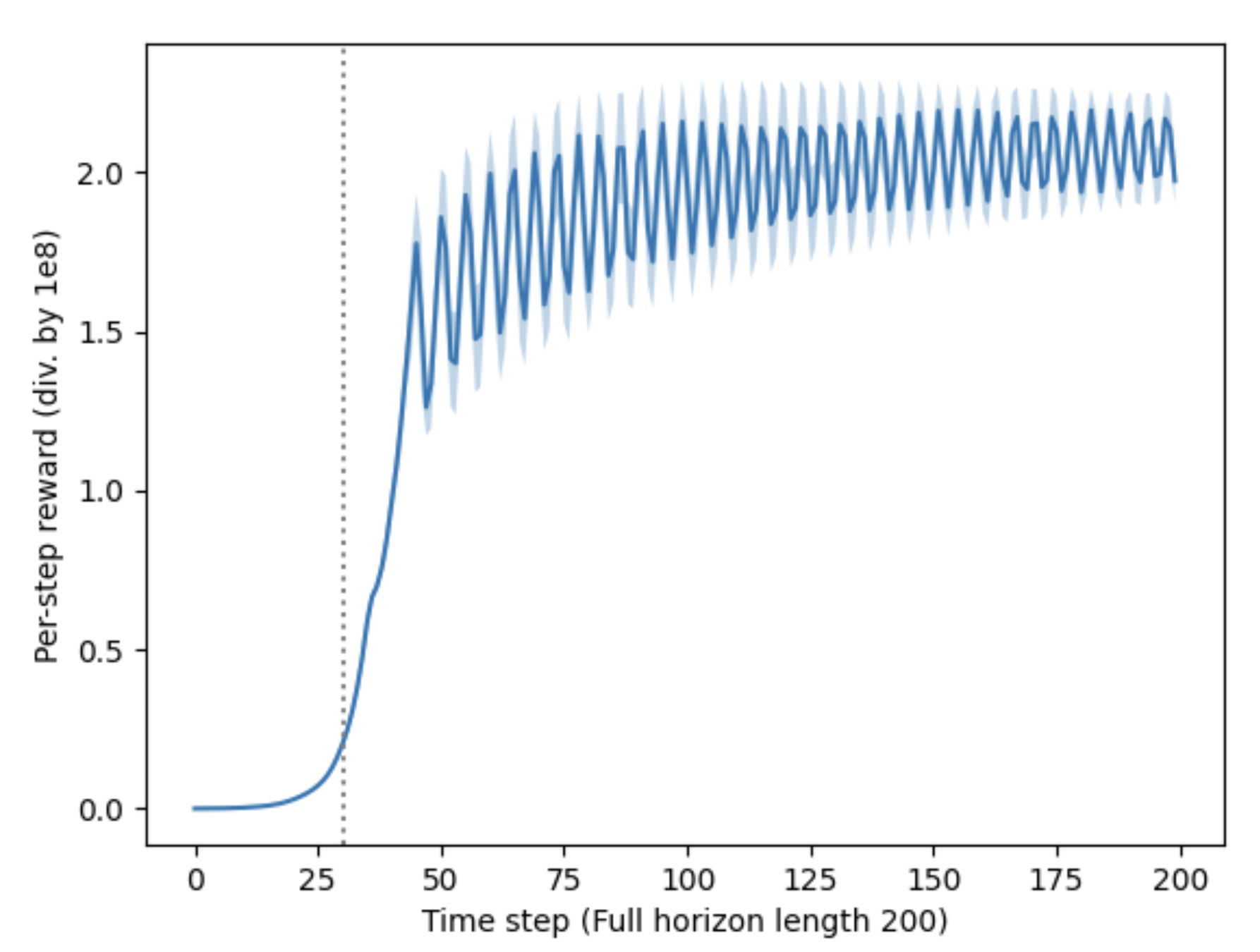}}
\caption{Example of a difficult-to-predict short-term signal in HIV treatment. Initially the policy accumulates very low rewards but eventually achieves a high return. Using only the last observed reward to extrapolate the future outcome is sub-optimal in this case as this significantly underestimates the long-term value (predicted: 40.6). On the other hand SLEV predicts the return to be 295.9 which is closer to the true value 313.7}
\label{fig:hiv_ex1}
\end{center}
\vskip -0.2in
\end{figure}

Fig \ref{fig:hiv_ex1} from HIV treatment shows the per-step reward sequence under some policy $\pi$ where the initial reward observations are very low but the final outcome value is very high (313.7). SLEV predicts the long-term value to be 295.9, and FQE predicts 201.7, but all other methods significantly underestimate (average reward extrapolation predicts 5.9, and last reward extrapolation predicts 40.6). These examples showcase how early reward signal may swap the ordering of the policies, but SLEV -- as long as prediction errors are reasonably small -- can still correctly rank the policies despite the misleading reward signal. 

\subsection{SLEV under distribution shift}
\begin{figure}[ht]
\vskip 0.2in
\begin{center}
\centerline{\includegraphics[scale=0.4]{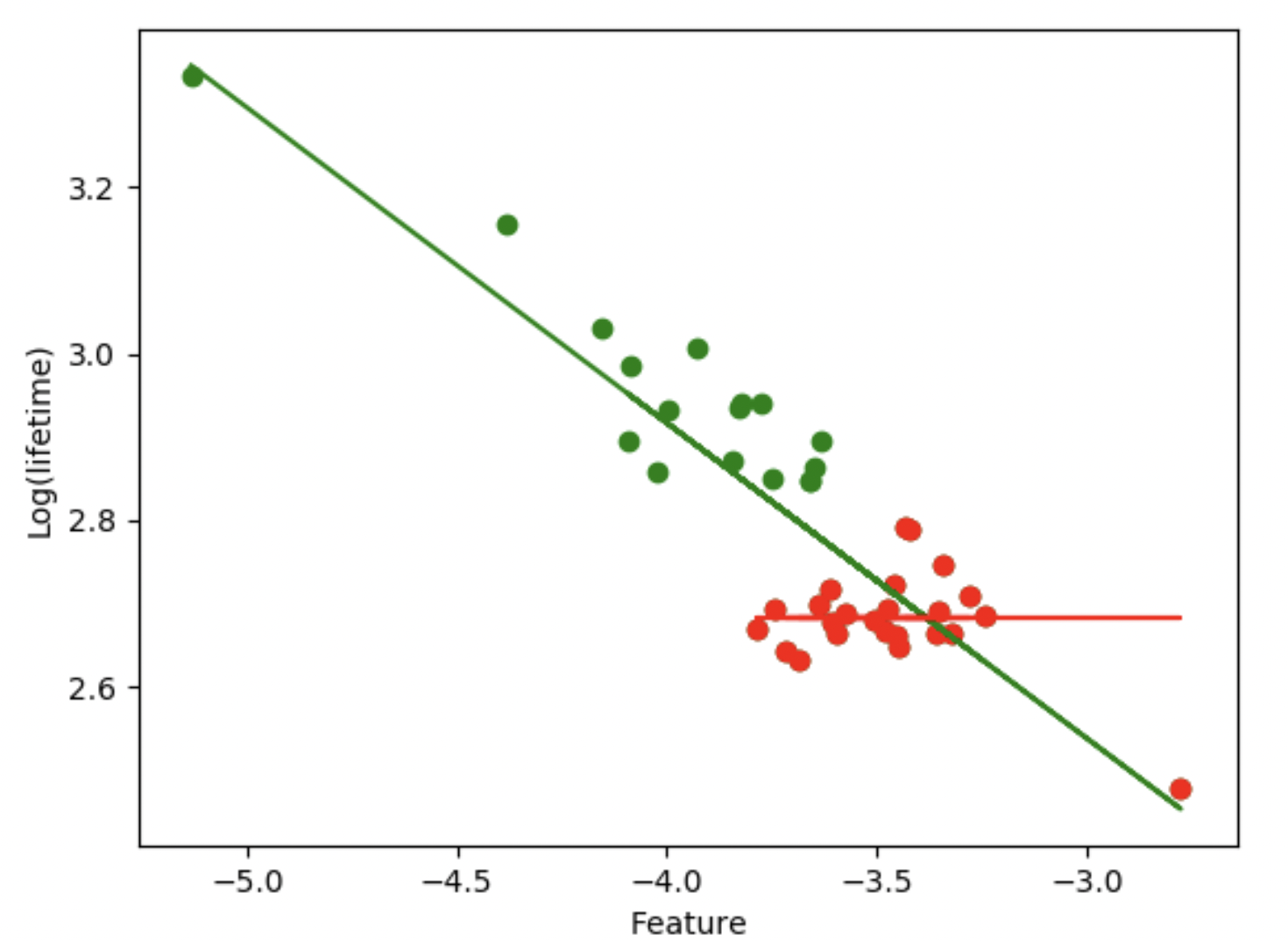}}
\caption{Scatter plot of 41 samples. Green uses all available points in $\mathcal D^{\text{train}}$, and red only uses a subset of 24.}
\label{fig:battery_failure}
\end{center}
\vskip -0.2in
\end{figure}

Fig \ref{fig:battery_failure} shows the best fitted lines to predict the battery life time (in log space) given the scalar feature ($\triangle Q_{5-4}$). The green line is fitted on the full 41 battery samples in the train data, and the red line is fitted only on the subset of 24 samples whose battery lifetimes are lower than 700. As suggested by this figure, SLEV achieves 86.7\% accuracy on the classification task when trained with the full data, but the accuracy drops to 26.5\% when trained only on the red points. This aligns with our understanding of regression-based methods that they are brittle to train-test distribution shift, and when the distribution mismatch is significant as in this case, the empirical risk minimizer on the train data generalizes poorly to the test samples. 

\subsection{Battery charging results}
\begin{table}
\label{table:battery_simplified_global_fcn}
\vskip 0.15in
\begin{center}
\begin{small}
\begin{sc}
\begin{tabular}{lcccr}
\toprule
Method & Threshold & Train size & Acc.\\
\midrule SLED ($\ell=5$) &  $< 700$ & 24 & 74.6 \\
Naive Bayes ($\ell=0$) &  & &  26.5 \\
$\triangle Q_{5-4}$  $(\ell=5)$ & & & 26.5  \\
\midrule SLED  &  N/A & 41  & 74.6 \\
Naive Bayes & (all samples incld) & &  26.5 \\
$\triangle Q_{5-4}$  & & & 86.7  \\
\bottomrule
\end{tabular}
\end{sc}
\end{small}
\end{center}
\vskip -0.1in
\end{table}

Table \ref{table:battery_simplified_global_fcn}
shows the result of predicting with SLED. Naive bayes predictor outputs the majority class from train data and cannot distinguish between optimal and sub-optimal battery charging policies. While SLEV performs well when the full train data is available, when the train data is filtered to only include examples with short lifetime of less than 700 cycles, the regression-based methods performance downgrade significantly. On the other hand, SLED performs well under both train distributions.

\section{Proof}\label{appendix:proof}
In order to state the theorem precisely, we first define some notations.

\subsection{Notation}
Let $P$ be training distribution and $Q$ be test distributions over $\mathcal X$.\footnote{Consider $P', Q'$ as joint distributions $P'(x, y), Q'(x, y)$ but under the assumption that $\mathbb P(y|x)$ is the same between train and test distributions, we can write $P'(x, y) = P(x)\mathbb P(y|x)$ and $Q'(x, y) = Q(x) \mathbb P(y|x)$. $P, Q$ denote the covariate only distributions, and the regression model $f$ is designed to learn $\mathbb P(y|x)$ from train data and be used for evaluating test data.} And $\mathcal D_P$, $\mathcal D_Q$ be empirical distributions (i.e,. datasets of samples from $P, Q$ respectively). Let $\mathcal D_P$ be split into $\mathcal D_1$ for estimating the density ratio and $\mathcal D_2$ for estimating regression model $f \in \mathcal F$. Let $m_1 = |\mathcal D_1|, m_2 = |\mathcal D_Q|, n = |\mathcal D_2|$ be the size of each dataset. 

Let $f$ be the regression model and $w \in \mathcal W$ be the density ratio function where $w(x) := \frac{Q(x)}{P(x)}$. For simplicity, we denote $x \in \mathcal X$ as inputs to the regression and $w$ and $y \in \mathcal Y$ as the regression target.

Let loss $l(f, x, y) := (f(x) - y)^2$ be the squared error. Note that since $J(\pi) \in [0, V_{\max}], l \leq V_{\max}^2$. 

We define test risk (= $\mathcal R_Q$) and training (empirical) risk $(= \hat {\mathcal R}_P)$ as: 

\begin{equation}
\mathcal R_Q(f) := \mathbb E_{(x, y) \sim Q} [l(f, x, y)]
\end{equation}

\begin{equation}
\hat {\mathcal R}_P(f) := \sum_{(x, y) \in \mathcal D_2} [l(f, x, y)]
\end{equation}

Weighted empirical risk by $w$ is: 
\begin{equation}
\hat {\mathcal R}_P(wf) := \sum_{(x, y) \in \mathcal D_2} w(x)[l(f, x, y)]
\end{equation}

\subsection{Assumption}
We require that $w$ be bounded, $\sup_{x} w(x), \hat w(x) \leq M$, which is satisfied if $\mathcal W := \{w: w \leq M\}$.

\subsection{Key Existing Results}

\begin{theorem}
Proposition 1 of \cite{cortes}. For  a single $f \in \mathcal F$, with probability at least $1 - \delta$,
\begin{equation}
|\mathcal R_Q(f) - \hat {\mathcal R}_P(wf)| \leq M V_{\max}^2 \sqrt{\frac{\log(2/\delta)}{2n}}
\end{equation} 
\end{theorem} Since their result is stated with $l \leq 1$, we modify to include $V_{\max}^2$ and importantly $w$ is the true density ratio and not the estimated one which is likely to add $||w - \hat w||_2$ error. To make the statement hold for all $f$ in finite class $\mathcal F$, union bound which only affects the log term. We combine this with \cite{kptofue} Theorem 1 to bound $|\mathcal R_Q(f) - \hat {\mathcal R}_P(\hat w f)|$ (i.e., test risk by the weighted empirical risk where the density ratio weights are also estimated and thus imperfect). 

We leave our final result in terms of $|| w - \hat w||_{1, P}$ since density ratio estimation is not the focus of this paper, and different bounds/techniques exist. For example, \cite{kptofue} bounds this in terms of $M$, lischiptiz-ness of $w$, the ``intrinsic" dimension of $\mathcal X$, as well as sample sizes $m_1, m_2$ used for density ratio estimation between $P$ and $Q$. \cite{nguyen} achieves $O(m^{-\frac{1}{\gamma + 2}})$ rate to bound 
 the error in the hellinger distance $h_P^2(\hat w, w)$ which is an upper bound on $||w - \hat w||_{1, P}$, where $\gamma$ describes the smoothness of functions in $\mathcal W$ and the train and test covariate datasets are assumed to be of the same size $m = m_1 = m_2$.

\subsection{Main Theorem}

\begin{theorem}
For $f \in \mathcal F$ (assume finite class $|\mathcal F| = F$) and density ratio estimate $\hat w$, with probability at least $1 - 4\delta$,
\begin{equation}
|\mathcal R_Q(f) - \hat{\mathcal R}_P(\hat w f)| \leq M V_{\max}^2 \sqrt{\frac{\log(2F/\delta)}{2n}} + V_{\max}^2 \left(\frac{M\sqrt{\log(2/\delta)}}{\sqrt{n}} + || w- \hat w||_{1, P} \right) 
\end{equation}
\end{theorem}

\paragraph{pf} For simplicity, consider a single hypothesis $f \in \mathcal F$. 
\begin{equation}
|\mathcal R_Q(f) - \hat{\mathcal R}_P(\hat w f)| = |\mathcal R_Q(f) - \hat {\mathcal R}_P(w f) + \hat {\mathcal R}_P(w f) - \hat {\mathcal R}_P(\hat w f)|
\end{equation}

\begin{equation}
\leq |\mathcal R_Q(f) - \hat {\mathcal R}_P(w f)| + |\hat {\mathcal R}_P(w f) - \hat {\mathcal R}_P(\hat w f)|
\end{equation}

Note the first term is bounded by C.1, so proceed with the second term.

\begin{equation}
|\hat {\mathcal R}_P(wf) - \hat {\mathcal R}_P (\hat w f)| = |\hat {\mathbb E}_{(x,y) \sim P} [(w(x) -  \hat w(x)) l (f, x,y)]|
\end{equation}

\begin{equation}
\leq \hat {\mathbb E}_{x \sim P} \left[V_{\max}^2 |w(x) - \hat w(x)|\right]
\end{equation}

Recall that $\mathbb E_{x \sim P} |w(x) - \hat w(x)|$ is bounded and $|w - \hat w|$ is upper bounded by $M$. By Hoeffding's the sample mean of $(w - \hat w)$ is also bounded as: 

\begin{equation}
\hat {\mathbb E}_{x \sim P}|w(x) - \hat w(x)| \leq M \sqrt{\frac{\log(2/\delta)}{n}} + ||w - \hat w||_{1, P}
\end{equation}

Combining eq (24), (26), and (27), and union bound the error probabilities, we obtain that:

\begin{equation}
(23) \leq M V_{\max}^2 \sqrt{\frac{\log(2F/\delta)}{2n}} + V_{\max}^2 \left(\frac{M\sqrt{\log(2/\delta)}}{\sqrt{n}} + || w- \hat w||_{1, P} \right) 
\end{equation} with probability at least $1 - 4\delta$. $\square$


\end{document}